\documentclass{article}

\usepackage{subcaption}
\usepackage[table]{xcolor}
\usepackage{hyperref}
\usepackage{enumitem}
\usepackage{arxiv}
\usepackage{natbib}
\usepackage[utf8]{inputenc} 
\usepackage[T1]{fontenc}    
\usepackage{hyperref}       
\usepackage{url}            
\usepackage{booktabs}       
\usepackage{amsfonts}       
\usepackage{nicefrac}       
\usepackage{microtype}      
\usepackage{lipsum}
\usepackage{graphicx}
\usepackage{academicons}
\usepackage{fancyhdr}
\setcitestyle{aysep={}}

\title{A Survey on Deep Learning-based Architectures\\for Semantic Segmentation on 2D images}

\author{
Irem~Ulku \\
    Department of Computer Engineering\\ Ankara University,\\
    Ankara, TURKEY\\
  \texttt{irem.ulku@ankara.edu.tr} \\
  ORCID-ID:0000-0003-4998-607X
   \And
Erdem~Akagündüz\\ 
    Graduate School of Informatics\\ Middle East Technical University,\\
    Ankara, TURKEY\\
  \texttt{akaerdem@metu.edu.tr} \\
  ORCID-ID:0000-0002-0792-7306 \\
  \emph{corresponding author}
}

\begin{document}
\maketitle
\thispagestyle{fancy}

\begin{abstract}
Semantic segmentation is the pixel-wise labelling of an image.
Boosted by the extraordinary ability of convolutional neural networks (CNN) in creating semantic, high level and hierarchical image features; several deep learning-based 2D semantic segmentation approaches have been proposed within the last decade. In this survey, we mainly focus on the recent scientific developments in semantic segmentation, specifically on deep learning-based methods using 2D images. We started with an analysis of the public image sets and leaderboards for 2D semantic segmentation, with an overview of the techniques employed in performance evaluation. In examining the evolution of the field, we chronologically categorised the approaches into three main periods, namely pre-and early deep learning era, the fully convolutional era, and the post-FCN era. We technically analysed the solutions put forward in terms of solving the fundamental problems of the field, such as fine-grained localisation and scale invariance. Before drawing our conclusions, we present a table of methods from all mentioned eras, with a summary of each approach that explains their contribution to the field. We conclude the survey by discussing the current challenges of the field and to what extent they have been solved.
\end{abstract}


\section{Introduction}

Semantic segmentation has recently become one of the fundamental problems, and accordingly, a hot topic for the fields of computer vision and machine learning. Assigning a separate class label to each pixel of an image is one of the important steps in building complex robotic systems such as driverless cars/drones, human-friendly robots, robot-assisted surgery, and intelligent military systems. Thus, it is no wonder that in addition to scientific institutions, industry-leading companies studying artificial intelligence are now summarily confronting this problem. 

The simplest problem definition for semantic segmentation is pixel-wise labelling. Because the problem is defined at the pixel level, finding only class labels that the scene includes is considered insufficient, but localising labels at the original image pixel resolution is also a fundamental goal. Depending on the context, class labels may change. For example, in a driverless car, the pixel labels may be \emph{human, road} and \emph{car} \citep{SiamEJY17} whereas for a medical system \citep{SahaTIP2018,Jiangmedical2017}, they could be \emph{cancer cells, muscle tissue, aorta wall} etc.  

The recent increase in interest in this topic has been undeniably caused by the extraordinary success seen with convolutional neural networks \citep{lecun1989generalization} (CNN) that have been brought to semantic segmentation. Understanding a scene at the semantic level has long been one of the main topics of computer vision, but it is only now that we have seen actual solutions to the problem.

In this paper, our primary motivation is to focus on the recent scientific developments in semantic segmentation, specifically on the evolution of deep learning-based methods using 2D images. The reason we narrowed down our survey to techniques that utilise only 2D visible imagery is that, in our opinion, the scale of the problem in the literature is so vast and widespread that it would be impractical to analyse and categorise all semantic segmentation modalities (such as 3D point clouds, hyper-spectral data, MRI, CT\footnote{We consider MRI and CT essentially as 3D volume data. Although individual MRI/CT slices are 2D, when doing semantic segmentation on these types of data, neighbourhood information in all three dimensions are utilised. For this reason, medical applications are excluded from this survey.} etc.) found in journal articles to any degree of detail. In addition to analysing the techniques which make semantic segmentation possible and accurate, we also examine the most popular image sets created for this problem. Additionally, we review the performance measures used for evaluating the success of semantic segmentation. Most importantly, we propose a taxonomy of methods, together with a technical evolution of them, which we believe is novel in the sense that it provides insight to the existing deficiencies and suggests future directions for the field.

The remainder of the paper is organised as follows: in the following subsection, we refer to other survey studies on the subject and underline our contribution. Section 2 presents information about the different image sets, the challenges, and how to measure the performance of semantic segmentation. Starting with Section 3, we chronologically scrutinise semantic segmentation methods under three main titles, hence in three separate sections. Section 3 covers the methods of pre- and early deep convolutional neural networks era. Section 4 provides details on the fully convolutional neural networks, which we consider a milestone for the semantic segmentation literature. Section 5 covers the state-of-the-art methods on the problem and provides details on both the architectural details and the success of these methods. Before finally concluding the paper in Section 7, Section 6 provides a future scope and potential directions for the field.

\subsection{Surveys on Semantic Segmentation}
Very recently, driven by both academia and industry, the rapid increase of interest in semantic segmentation has inevitably led to a number of survey studies being published \citep{thoma2016,Ahmad_2017,Jiangmedical2017,SiamEJY17,Garcia2017,saffar2018semantic,YU201882,Guo2018,LATEEF2019,minaee2020image}.

Some of these surveys focus on a specific problem, such as comparing semantic segmentation approaches for horizon/skyline detection \citep{Ahmad_2017}, whilst others deal with relatively broader problems related to industrial challenges, such as semantic segmentation for driverless cars \citep{SiamEJY17}  or medical systems \citep{Jiangmedical2017}. These studies are useful if working on the same specific problem, but they lack an overarching vision that may `technically' contribute to the future directions of the field.

Another group \citep{thoma2016,saffar2018semantic,YU201882,Guo2018} of survey studies on semantic segmentation have provided a general overview of the subject, but they lack the necessary depth of analysis regarding deep learning-based methods. Whilst semantic segmentation was studied for two decades prior to deep learning, actual contribution to the field has only been achieved very recently, particularly following a revolutionary paper on fully convolutional networks (FCN) \citep{Shelhamer2017} (which has also been thoroughly analysed in this paper). It could be said that most state-of-the-art studies are in fact extensions of that same \citep{Shelhamer2017} study. For this reason, without scrupulous analysis of FCNs and the direction of the subsequent papers, survey studies will lack the necessary academic rigour in examining semantic segmentation using deep learning.


There are recent reviews of deep semantic segmentation, for example by \citep{Garcia2017} and \citep{minaee2020image}, which provide a comprehensive survey on the subject. These survey studies cover almost all the popular semantic segmentation image sets and methods, and for all modalities such as 2D, RGB, 2.5D, RGB-D, and 3D data. Although they are inclusive in the sense that most related material on deep semantic segmentation is included, the categorisation of the methods is coarse, since the surveys attempt to cover almost everything umbrellaed under the topic of semantic segmentation literature. 

A detailed categorisation of the subject was provided in \citep{LATEEF2019}. Although this survey provides important details on the subcategories that cover almost all approaches in the field, discussions on how the proposed techniques are chronologically correlated are left out of their scope. Recent deep learning studies on semantic segmentation follow a number of fundamental directions and labour with tackling the varied corresponding issues. In this survey paper, we define and describe these new challenges, and present the chronological evolution of the techniques of all the studies within this proposed context. We believe our attempt to understand the evolution of semantic segmentation architectures is the main contribution of the paper. We provide a table of these related methods, and explain them briefly one after another in chronological order, with their metric performance and computational efficiency. This way, we believe that readers will better understand the evolution, current state-of-the-art, as well as the future directions seen for 2D semantic segmentation.

\section{Image Sets, Challenges and Performance Evaluation}

\subsection{Image Sets and Challenges}
\label{datasets}
The level of success for any machine-learning application is undoubtedly determined by the quality and the depth of the data being used for training. When it comes to deep learning, data is even more important since most systems are termed end-to-end; thus, even the features are determined \emph{by} the data, not \emph{for} the data. Therefore, data is no longer the object but becomes the actual subject in the case of deep learning.

In this section, we scrutinise the most popular large-scale 2D image sets that have been utilised for the semantic segmentation problem. The image sets were categorised into two main branches, namely general-purpose image sets, with generic class labels including almost every type of object or background, and also urban street image sets, which include class labels such as car and person, and are generally created for the training of driverless car systems. There are many other unresolved 2D semantic segmentation problem domains such as medical imaging, satellite imagery, or infrared imagery. However, urban street image is currently driving scientific development in the field because they attract more attention from the industry. Therefore, very large-scale image sets and challenges with crowded leaderboards exist, yet, only specifically for industrial users. Scientific interest for depth-based semantic segmentation is growing rapidly; however, as mentioned in the Introduction, we have excluded depth-based and 3D-based segmentation datasets from the current study in order to focus with sufficient detail on the novel categorisation of recent techniques pertinent to 2D semantic segmentation.

\subsubsection{General Purpose Semantic Segmentation Image Sets}
\begin{itemize}
\item PASCAL Visual Object Classes (VOC) \citep{Everingham2010}: This image set includes image annotations not only for semantic segmentation, but for also classification, detection, action classification, and person layout tasks. The image set and annotations are regularly updated and the leaderboard of the challenge is public\footnote{\url{ http://host.robots.ox.ac.uk:8080/leaderboard/main\_bootstrap.php}} (with more than 140 submissions just for the segmentation challenge alone). It is the most popular among the semantic segmentation challenges and is still active following its initial release in 2005. The PASCAL VOC semantic segmentation challenge image set includes 20 foreground object classes and one background class. The original data consisted of 1,464 images for the purposes of training, plus 1,449 images for validation. The 1,456 test images are kept private for the challenge. The image set includes all types of indoor and outdoor images and is generic across all categories.

The PASCAL VOC image set has a number of extension image sets, the most popular among these are PASCAL Context \citep{mottaghi_cvpr14} and PASCAL Parts \citep{chen_cvpr14}. The first \citep{mottaghi_cvpr14} is a set of additional annotations for PASCAL VOC 2010, which goes beyond the original PASCAL semantic segmentation task by providing annotations for the whole scene. The statistics section contains a full list of more than 400 labels (compared to the original 21 labels). The second \citep{chen_cvpr14} is also a set of additional annotations for PASCAL VOC 2010. It provides segmentation masks for each body part of the object, such as the separately labelled limbs and body of an animal. For these extensions, the training and validation set contains 10,103 images, while the test set contains 9,637 images. There are other extensions to PASCAL VOC using other functional annotations such as the Semantic Parts (PASParts) \citep{Wangiccv2015} image set and the Semantic Boundaries Dataset (SBD) \citep{Hariharan2011}. For example, PASParts \citep{Wangiccv2015} additionally provides `instance’ labels such as two instances of an object within an image are labelled separately, rather than using a single class label. However, unlike the former two additional extensions \citep{chen_cvpr14,mottaghi_cvpr14}, these further extensions \citep{Wangiccv2015,Hariharan2011} have proven less popular as their challenges have attracted much less attention in state-of-the-art semantic segmentation studies; thus, their leaderboards are less crowded. In Figure \ref{PascalParts}, a sample object, parts and instance segmentation are depicted.

\begin{figure*}[t]
\centering
\includegraphics*[clip=false,width=1\textwidth]{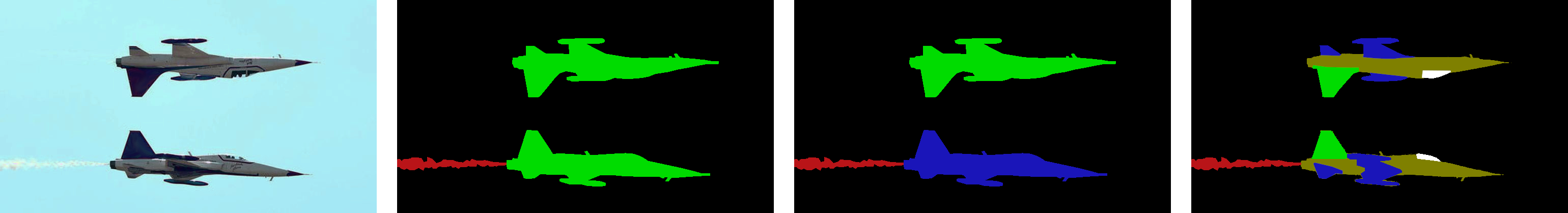}
\caption{A sample image and its annotation for object, instance and parts segmentations separately, from left to right.}
 \label{PascalParts}
\end{figure*} 

\item Common Objects in Context (COCO) \citep{lin2014microsoft}: With 200K labelled images, 1.5 million object instances, and 80 object categories, COCO is a very large scale object detection, semantic segmentation, and captioning image set, including almost every possible types of scene. COCO provides challenges not only at the instance-level and pixel-level (which they refer to as \emph{stuff}) semantic segmentation, but also introduces a novel task, namely that of \emph{panoptic} segmentation \citep{Kirillov18}, which aims at unifying instance-level and pixel-level segmentation tasks. Their leaderboards\footnote{\url{http://cocodataset.org}}  are relatively less crowded because of the scale of the data. On the other hand, for the same reason, their challenges are assessed only by the most ambitious scientific and industrial groups, and thus are considered as the state-of-the-art in their leaderboards. Due to its extensive volume, most studies partially use this image set to pre-train or fine-tune their model, before submitting to other challenges such as PASCAL VOC 2012. 

\item ADE20K dataset \citep{Zhou2019}: ADE20K contains more than 20K scene-centric images with objects and object parts annotations. Similarly to PASCAL VOC, there is a public leaderboard\footnote{\url{http://sceneparsing.csail.mit.edu/}} and the benchmark is divided into 20K images for training, 2K images for validation, and another batch of held-out images for testing. The samples in the dataset have varying resolutions (average image size being 1.3M pixels), which can be up to 2400$\times$1800 pixels. There are total of150 semantic categories included for evaluation.

\item Other General Purpose Semantic Segmentation Image Sets: Although less popular than either PASCAL VOC or COCO, there are also some other image sets in the same domain. Introduced by \citep{Prest2012}, YouTube-Objects is a set of low-resolution (480$\times$360) video clips with more than 10k pixel-wise annotated frames. 

Similarly, SIFT-flow \citep{SIFTFlow} is another low-resolution (256$\times$256) semantic segmentation image set with 33 class labels for a total of 2,688 images. These and other relatively primitive image sets have been mostly abandoned in the semantic segmentation literature due to their limited resolution and low volume.

\end{itemize}

\subsubsection{Urban Street Semantic Segmentation Image Sets}

\begin{itemize}

\item Cityscapes \citep{cordts2016cityscapes}: This is a largescale image set with a focus on the semantic understanding of urban street scenes. It contains annotations for high-resolution images from 50 different cities, taken at different hours of the day and from all seasons of the year, and also with varying backgrounds and scene layouts. The annotations are carried out at two quality levels: fine for 5,000 images and course for 20,000 images. There are 30 different class labels, some of which also have instance annotations (vehicles, people, riders etc.). Consequently, there are two challenges with separate public leaderboards\footnote{\url{https://www.cityscapes-dataset.com/benchmarks/}}: one for pixel-level semantic segmentation, and a second for instance-level semantic segmentation. There are more than 100 entries to the challenge, making it the most popular regarding semantic segmentation of urban street scenes.

\item Other Urban Street Semantic Segmentation Image Sets: There are a number of alternative image sets for urban street semantic segmentation, such as Cam-Vid \citep{Brostow2009SemanticOC}, KITTI \citep{Geiger2013}, SYNTHIA \citep{RosCVPR16}, and IDD \citep{IDD-Dataset}. These are generally overshadowed by the Cityscapes image set \citep{cordts2016cityscapes} for several reasons. Principally, their scale is relatively low. Only the SYNTHIA image set \citep{RosCVPR16} can be considered as largescale (with more than 13k annotated images); however, it is an artificially generated image set, and this is considered a major limitation for security-critical systems like driverless cars.
\end{itemize}
	
\subsubsection{Small-scale and Imbalanced Image Sets}
In addition to the aforementioned large-scale image sets of different categories, there are several image sets with insufficient scale or strong imbalance such that, when applied to deep learning-based semantic segmentation models, high-level segmentation accuracies cannot be directly obtained. Most public challenges on semantic segmentation include sets of this nature such as the DSTL or RIT-18 \citep{dstl,kemker2018algorithms}, just to name a few. Because of the overwhelming numbers of these types of sets, we chose to include only the details of the large-scale sets that attract the utmost attention from the field. 

Nonetheless, being able to train a model that performs well on small-scale or imbalanced data is a correlated problem to ours. Besides conventional deep learning techniques such as transfer learning or data augmentation; the problem of insufficient or imbalanced data can be attacked by using specially designed deep learning architectures such as some optimized convolution layer types (\citet{Chen18,KaimingHeSPP}, etc.) and others that we cover in this survey paper. What is more, there are recent studies that focus on the specific problem of utilizing insufficient sets for the problem of deep learning-based semantic segmentation \citep{Xia_2019}. Although we acknowledge this problem as fundamental for the semantic segmentation field, we leave the discussions on techniques to handle small-scale or imbalanced sets for semantic segmentation, beyond the scope of this survey paper.

\subsection{Performance Evaluation}
There are two main criteria in evaluating the performance of semantics segmentation: accuracy, or in other words, the success of an algorithm; and computation complexity in terms of speed and memory requirements. In this section, we analyse these two criteria separately.

\subsubsection{Accuracy}
Measuring the performance of segmentation can be complicated, mainly because there are two distinct values to measure. The first is classification, which is simply determining the pixel-wise class labels; and the second is localisation, or finding the correct set of pixels that enclose the object. Different metrics can be found in the literature to measure one or both of these values. The following is a brief explanation of the principal measures most commonly used in evaluating semantic segmentation performance. 

\begin{itemize}
    \item\emph{ROC-AUC}: ROC stands for the Receiver-Operator Characteristic curve, which summarises the trade-off between true positive rate and false-positive rate for a predictive model using different probability thresholds; whereas AUC stands for the area under this curve, which is 1 at maximum. This tool is useful in interpreting binary classification problems and is appropriate when observations are balanced between classes. However, since most semantic segmentation image sets \citep{Everingham2010,mottaghi_cvpr14,chen_cvpr14,Wangiccv2015,Hariharan2011,lin2014microsoft,cordts2016cityscapes} are not balanced between the classes, this metric is no longer used by the most popular challenges.

    \item\emph{Pixel Accuracy}: Also known as \emph{global accuracy} \citep{BadrinarayananK15}, pixel accuracy (PA) is a very simple metric which calculates the ratio between the amount of properly classified pixels and their total number. Mean pixel accuracy (mPA), is a version of this metric which computes the ratio of correct pixels on a per-class basis. mPA is also referred to as \emph{class average accuracy} \citep{BadrinarayananK15}. 

    \begin{equation}\label{pa}
        PA=\frac{\sum_{j=1}^{k}{n_{jj}}}{\sum_{j=1}^{k}{t_{j}}}, \qquad mPA=\frac{1}{k}\sum_{j=1}^{k}\frac{n_{jj}}{t_{j}}
    \end{equation}     
     
    where $n_{jj}$ is the total number of pixels both classified and labelled as class \emph{j}. In other words, $n_{jj}$ corresponds to the total number of \textit{True Positives} for class \emph{j}. $t_{j}$  is the total number of pixels labelled as class \emph{j}.

    \item\emph{Intersection over Union} (IoU): Also known as the Jaccard Index, IoU is a statistic used for comparing the similarity and diversity of sample sets. In semantics segmentation, it is the ratio of the intersection of the pixel-wise classification results with the ground truth, to their union.
    
    \begin{equation}\label{iu}
    IoU=\frac{\sum_{j=1}^{k}{n_{jj}}}{\sum_{j=1}^{k}({n_{ij}+n_{ji}+n_{jj}})}, \qquad i \neq j
    \end{equation}
    
    where, $n_{ij}$ is the number of pixels which are labelled as class \emph{i}, but classified as class \emph{j}. In other words, they are  \textit{False Positives} (false alarms) for class \emph{j}. Similarly, ${n_{ji}}$, the total number of pixels labelled as class \emph{j}, but classified as class \emph{i} are the \textit{False Negatives} (misses) for class \emph{j}.
            
    Two extended versions of IoU are also widely in use:

    $\circ$\emph{ Mean Intersection over Union} (mIoU): mIoU is the class-averaged IoU, as in (\ref{miou}).
    
    \begin{equation}\label{miu}
        mIoU=\frac{1}{k}\sum_{j=1}^{k}\frac{n_{jj}}{n_{ij}+n_{ji}+n_{jj}}, \qquad i \neq j
    \label{miou}    
    \end{equation}
    
    $\circ$\emph{ Frequency-weighted IoU} (FwIoU): This is an improved version of MIoU that weighs each class importance depending on appearance frequency by using $t_{j}$ (the total number of pixels labelled as class \emph{j}, as also defined in  (\ref{pa})). The formula of FwIoU is given in (\ref{fiu}):
    
    \begin{equation}\label{fiu}
        FwIoU=\frac{1}{\sum_{j=1}^{k}t_{j}}\sum_{j=1}^{k}{t_{j}\frac{n_{jj}}{n_{ij}+n_{ji}+n_{jj}}}, \qquad i \neq j
    \end{equation}
    
     IoU and its extensions, compute the ratio of true positives (hits) to the sum of false positives (false alarms), false negatives (misses) and true positives (hits). Thereby, the IoU measure is more informative when compared to pixel accuracy simply because it takes false alarms into consideration, whereas PA does not. However, since false alarms and misses are summed up in the denominator, the significance between them is not measured by this metric, which is considered its primary drawback. In addition, IoU only measures the number of pixels correctly labelled without considering how accurate the segmentation boundaries are.

     \item\emph{Precision-Recall Curve (PRC)-based metrics}: Precision (ratio of hits over a summation of hits and false alarms) and recall (ratio of hits over a summation of hits and misses) are the two axes of the PRC used to depict the trade-off between precision and recall, under a varying threshold for the task of binary classification. PRC is very similar to ROC. However, PRC is more powerful in discriminating the effects between the false positives (alarms) and false negatives (misses). That is predominantly why PRC-based metrics are commonly used for evaluating the performance of semantic segmentation. The formula for Precision (also called Specificity) and Recall (also called Sensitivity) for a given class \emph{j}, are provided in (\ref{prerec}):
    
    \begin{equation}\label{prerec}
       Prec.=\frac{n{_{jj}}}{n_{ij}+n_{jj}}, \quad
        Recall=\frac{n_{jj}}{n_{ji}+n_{jj}},  i \neq j
    \end{equation}

    There are three main PRC-based metrics:
    
   $\circ$\emph{ F}$_{score}$:  Also known as the ‘\emph{dice coefficient}’, this measure is the harmonic mean of the precision and recall for a given threshold. It is a normalised measure of similarity, and ranges between 0 and 1 (Please see (\ref{f})).
    
    \begin{equation}\label{f}
        F_{score}=2 \times \frac{Precision\times Recall}{Precision+Recall}
    \end{equation}
     
    $\circ$\emph{ PRC-AuC}: This is similar to the ROC-AUC metric. It is simply the area under the PRC. This metric refers to information about the precision-recall trade-off for different thresholds, but not the \emph{shape} of the PR curve.
    
    $\circ$\emph{ Average Precision} (AP): This metric is a single value that summarises both the shape and the AUC of PRC. In order to calculate AP, using the PRC, for uniformly sampled recall values (e.g., 0.0, 0.1, 0.2, ..., 1.0), precision values are recorded. The average of these precision values is referred to as the average precision. This is the most commonly used single value metric for semantic segmentation. Similarly, mean average precision (mAP) is the mean of the AP values, calculated on a per-class basis.
    
       \item\emph{Hausdorff Distance} (HD): Hausdorff Distance is used incorporating the longest distance between classified and labelled pixels as an indicator of the largest segmentation error \citep{karimi2019reducing, jadon2020survey}, with the aim of tracking the performance of a semantic segmentation model. The unidirectional HDs as $hd(X,Y)$ and $hd(Y,X)$ are presented in (\ref{hdu1}) and (\ref{hdu2}), respectively.
   
    \begin{equation}\label{hdu1}
    hd\left ( X, Y \right )= \max_{{x\epsilon X}}\min_{{y\epsilon Y}}\left \| x-y \right \|_{2},
    \end{equation}

    \begin{equation}\label{hdu2}
    hd\left ( Y, X \right )= \max_{{y\epsilon Y}}\min_{{x\epsilon X}}\left \| x-y \right \|_{2}.
    \end{equation}

  where, $X$ and $Y$ are the pixel sets. The $x$ is the pixel in the segmented counter $X$ and $y$ is the pixel in the target counter $Y$ \citep{huang2020channel}. The bidirectional HD between these sets is shown in (\ref{hdb}), where the Euclidean distance is employed for (\ref{hdu1}), (\ref{hdu2}) and (\ref{hdb}).

    \begin{equation}\label{hdb}
    HD\left ( X,Y \right )= max\left ( hd\left ( X,Y \right ),hd\left ( Y,X \right ) \right ).
    \end{equation}

\end{itemize}

IoU and its variants, along with AP, are the most commonly used accuracy evaluation metrics in the most popular semantic segmentation challenges \citep{Everingham2010,mottaghi_cvpr14,chen_cvpr14,Wangiccv2015,Hariharan2011,lin2014microsoft,cordts2016cityscapes}.

\subsubsection{Computational Complexity}
The burden of computation is evaluated using two main metrics: how fast the algorithm completes, and how much computational memory is demanded. 

\begin{itemize}

\item\emph{Execution time}: This is measured as the whole processing time, starting from the instant a single image is introduced to the system/algorithm right through until the pixel-wise semantic segmentation results are obtained. The performance of this metric significantly depends on the hardware utilised. Thus, for an algorithm, any execution time metric should be accompanied by a thorough description of the hardware used. There are notations such as Big-O, which provide a complexity measure independent of the implementation domain. However, these notations are highly theoretical and are predominantly not preferred for extremely complex algorithms such as deep semantic segmentation as they are simple and largely inaccurate.
    
For a deep learning-based algorithm, the offline (i.e., training) and online (i.e., testing) operation may last for considerably different time intervals. Technically, the execution time refers only to the online operation or, academically speaking, the test duration for a single image. Although this metric is extremely important for industrial applications, academic studies refrain from publishing exact execution times, and none of the aforementioned challenges was found to have provided this metric. A recent study, \citep{zhao2018icnet} provided a 2D histogram of Accuracy (MIoU\%) vs frames-per-second, in which some of the state-of-the-art methods with open source codes (including their proposed structure, namely image cascade network – ICNet), were benchmarked using the Cityscapes \citep{cordts2016cityscapes}  image set.

\item\emph{Memory Usage}: Memory usage is specifically important when semantic segmentation is utilised in limited performance devices such as smartphones, digital cameras, or when the requirements of the system are extremely restrictive. The prime examples of these would be military systems or security-critical systems such as self-driving cars. 
  
The usage of memory for a complex algorithm like semantic segmentation may change drastically during operation. That is why a common metric for this purpose is \emph{peak memory usage}, which is simply the maximum memory required for the entire segmentation operation for a single image. The metric may apply to computer (data) memory or GPU memory depending on the hardware design.
    
Although critical for industrial applications, this metric is not usually made available for any of the aforementioned challenges.

\end{itemize}

Computational efficiency is a very important aspect of any algorithm that is to be implemented on a real system. A comparative assessment of the speed and capacity of various semantic segmentation algorithms is a challenging task. Although most state-of-the-art algorithms are available with open-source codes, benchmarking all of them, with their optimal hyper-parameters, seems implausible. 
For this purpose, we provide an inductive way of comparing the computational efficiencies of methods in the following sections. In Table \ref{MethodsTable}, we categorise methods into mainly four levels of computational efficiency and discuss our categorisation related to the architectural design of a given method. This table also provides a chronological evolution of the semantic segmentation methods in the literature. 

\section{Before Fully Convolutional Networks}
As mentioned in the Introduction, the utilisation of FCNs is a breaking point for semantic segmentation literature. Efforts on semantic segmentation literature prior to FCNs \citep{Shelhamer2017} can be analysed in two separate branches, as pre-deep learning and early deep learning approaches. In this section, we briefly discuss both sets of approaches.

\subsection{Pre-Deep Learning Approaches}
The differentiating factor between conventional image segmentation and semantic segmentation is the utilisation of semantic features in the process. Conventional methods for image segmentation such as thresholding, clustering, and region growing, etc. (\emph{please see} \citep{ZAITOUN2015} \emph{for a survey on conventional image segmentation techniques}) utilise handcrafted low-level features (i.e., edges, blobs) to locate object boundaries in images. Thus, in situations where the semantic information of an image is necessary for pixel-wise segmentation, such as in similar objects occluding each other, these methods usually return a poor performance.

Regarding semantic segmentation efforts prior to DCNNs becoming popular, a wide variety of approaches \citep{HeNIPS2008,UlusoyCVPR05,LadickICCV2009,Bjorn2013,Montillo2011,Ravi2016,Vezhnevets2011,Shotton2008,Yao2012,XiaoICCV209,Micuslik2009,PylonModel2011,krahenbuhl2011} utilised graphical models, such as Markov Random Fields (MRF), Conditional Random Fields (CRF) or forest-based (or sometimes referred to as ‘holistic’) methods, in order to find scene labels at the pixel level. The main idea was to find an inference by observing the dependencies between neighbouring pixels. In other words, these methods modelled the semantics of the image as a kind of prior information among adjacent pixels. Thanks to deep learning, today we know that image semantics require abstract exploitation of largescale data. Initially, graph-based approaches were thought to have this potential. The so-called ‘super-pixelisation’, which is usually the term applied in these studies, was a process of modelling abstract regions. However, a practical and feasible implementation for largescale data processing was never achieved for these methods, while it was accomplished for DCNNs, first by \citep{AlexNet2012} and then in many other studies.

Another group of studies, sometimes referred to as the ‘Layered models’ \citep{yang2012,arbelaez2012semantic,LadickyECCV2010}, used a composition of pre-trained and separate object detectors so as to extract the semantic information from the image. Because the individual object detectors failed to classify regions properly, or because the methods were limited by the finite number of object classes provided by the ‘hand-selected’ bank of detectors in general, their performance was seen as relatively low compared to today’s state-of-the-art methods.

Although the aforementioned methods of the pre-deep learning era are no longer preferred as segmentation methods, some of the graphical models, especially CRFs, are currently being utilised by the state-of-the-art methods as post-processing (refinement) layers, with the purpose of improving the semantic segmentation performance, the details of which are discussed in the following section.

\begin{figure*}[t]
\centering
\begin{subfigure}{4.8cm}	
    \centering
	\includegraphics*[trim=0 0 0 0,width=1.0\textwidth]{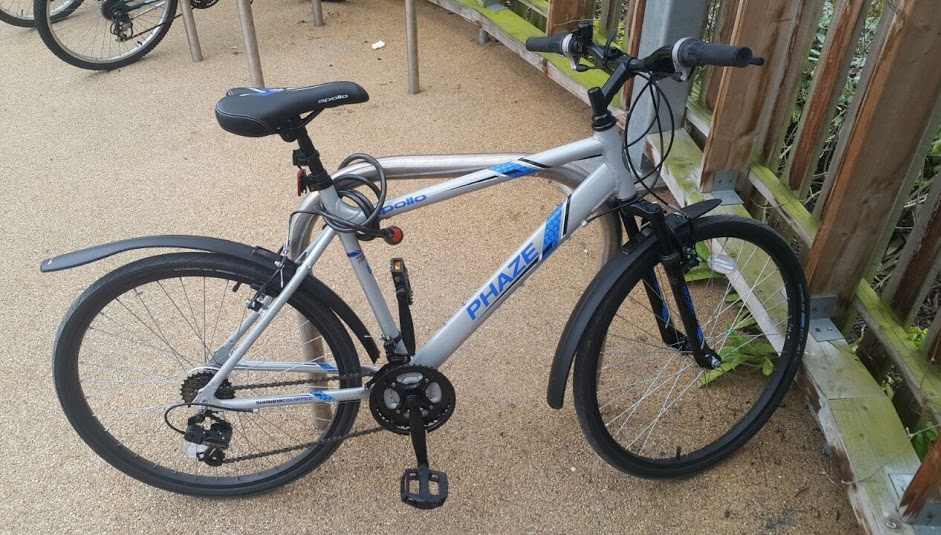}
	\caption{Input Image}	
\end{subfigure}
\begin{subfigure}{4.8cm}
    \centering
	\includegraphics*[trim=0 0 0 0,width=1.0\textwidth]{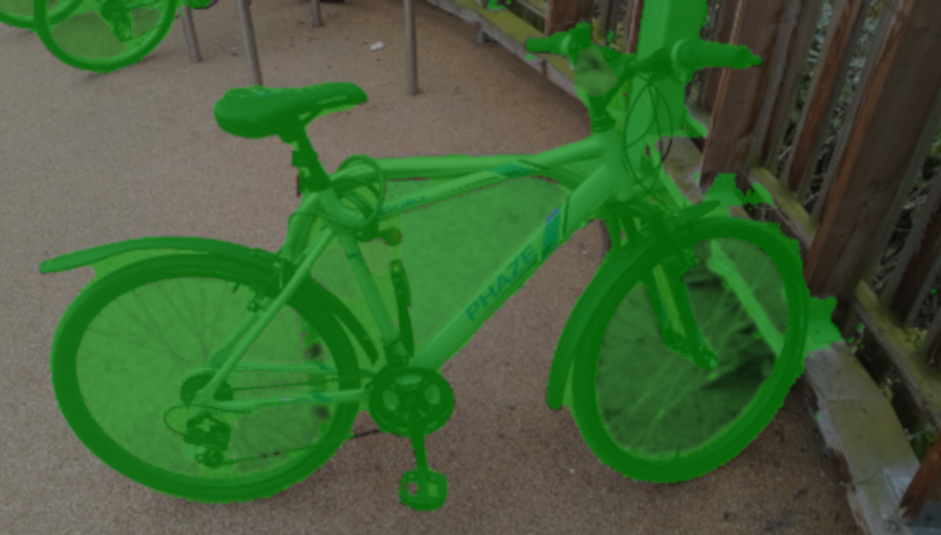}
	\caption{Segmented Image}	
\end{subfigure}
\begin{subfigure}{4.8cm}
    \centering
	\includegraphics*[trim=0 0 0 0,width=1.0\textwidth]{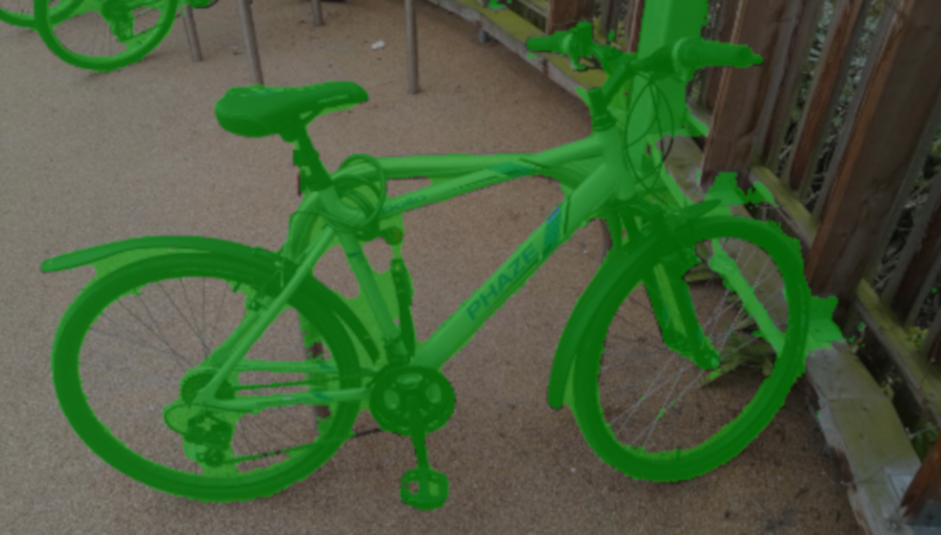}
	\caption{Refined Result}	 
\end{subfigure}
\caption{Effect of using graphical model-based refinement on segmentation results.}
 \label{CRFRefinement}
\end{figure*}

\subsubsection{Refinement Methods}
Deep neural networks are powerful in extracting abstract local features. However, they lack the capability to utilise global context information, and accordingly cannot model interactions between adjacent pixel predictions \citep{Marvin2018}. On the other hand, the popular segmentation methods of the pre-deep learning era, the graphical models, are highly suited to this sort of task. That is why they are currently being used as a refinement layer on many DCNN-based semantic segmentation architectures.

As also mentioned in the previous section, the idea behind using graphical models for segmentation is finding an inference by observing the low-level relations between neighbouring pixels. In Figure \ref{CRFRefinement}, the effect of using a graphical model-based refinement on segmentation results can be seen. The classifier (see Figure \ref{CRFRefinement}.b) cannot correctly segment pixels where different class labels are adjacent. In this example, a CRF-based refinement \citep{krahenbuhl2011} is applied to improve the pixel-wise segmentation results. CRF-based methods are widely used for the refinement of deep semantic segmentation methods, although some alternative graphical model-based refinement methods also exist in the literature \citep{Liu2015Semantic,pmlr-v78-zuo17a}. 

CRFs \citep{lafferty2001} are a type of discriminative undirected probabilistic graphical model. They are used to encode known relationships between observations and to construct consistent interpretations. Their usage as a refinement layer comes from the fact that, unlike a discrete classifier, which does not consider the similarity of adjacent pixels, a CRF can utilise this information. The main advantage of CRFs over other graphical models (such as Hidden Markov Models) is their conditional nature and their ability to avoid the problem of label bias \citep{lafferty2001}. Even though a considerable number of methods (see Table \ref{MethodsTable}) utilise CRFs for refinement, these models started to lose popularity in relatively recent approaches because they are notoriously slow and very difficult to optimise \citep{Marvin2018}.

\subsection{Early Deep Learning Approaches}
\label{pre-early-deep}

Before FCNs first appeared in 2014\footnote{FCN \citep{Shelhamer2017} ] was officially published in 2017. However, the same group first shared the idea online as pre-printed literature in 2014  \citep{Long2014Arxiv}.}, the initial few years of deep convolutional networks saw a growing interest in the idea of utilising the newly discovered deep features for semantic segmentation \citep{Ning2005,Ganin2014,Ciresan2012,Farabet2013,Hariharan2014,Pinheiro2014}. The very first approaches, which were published prior to the proposal of a Rectified Linear Unit (ReLU) layer \citep{AlexNet2012}, used activation functions such as \emph{tanh} \citep{Ning2005} (or similar continuous functions), which can be computationally difficult to differentiate. Thus, training such systems were not considered to be computation-friendly, or even feasible for largescale data.

However, the first mature approaches were just simple attempts to convert classification networks such Alex-Net and VGG to segmentation networks by fine-tuning the fully connected layers \citep{Ning2005,Ganin2014,Ciresan2012}. They suffered from the overfitting and time-consuming nature of their fully connected layers in the training phase. Moreover, the CNNs used were not sufficiently deep so as to create abstract features, which would relate to the semantics of the image.

There were a few early deep learning studies in which the researchers declined to use fully connected layers for their decision-making. However, they utilised different structures such as a recurrent architecture \citep{Pinheiro2014} or using labelling from a family of separately computed segmentations \citep{Farabet2013}. By proposing alternative solutions to fully connected layers, these early studies showed the first traces of the necessity for a structure like the FCN, and unsurprisingly they were succeeded by \citep{Shelhamer2017}.

Since their segmentation results were deemed to be unsatisfactory, these studies generally utilised a refinement process, either as a post-processing layer\citep{Ning2005,Ganin2014,Ciresan2012,Hariharan2014} or as an alternative architecture to fully connected decision layers \citep{Farabet2013,Pinheiro2014}. Refinement methods varied, such as Markov random fields \citep{Ning2005}, nearest neighbour-based approach \citep{Ganin2014}, the use of a calibration layer \citep{Ciresan2012}, using super-pixels \citep{Farabet2013,Hariharan2014}, or a recurrent network of plain CNNs \citep{Pinheiro2014}. Refinement layers, as discussed in the previous section, are still being utilised by post-FCN methods, with the purpose of increasing the pixel-wise labelling performance around regions where class intersections occur.

\section{Fully Convolutional Networks for Semantic Segmentation}
In \citep{Shelhamer2017}, the idea of dismantling fully connected layers from deep CNNs (DCNN) was proposed, and to imply this idea, the proposed architecture was named as ‘Fully Convolutional Networks’ (see Figure 3). The main objective was to create semantic segmentation networks by adapting classification networks such as AlexNet \citep{Krizhevsky2012}, VGG \citep{Simonyan15} , and GoogLeNet \citep{Szegedy2015} into fully convolutional networks, and then transferring their learnt representations by fine-tuning. The most widely used architectures obtained from the study \citep{Shelhamer2017} are known as ‘FCN-32s’, ‘FCN16s’, and ‘FCN8s’, which are all transfer-learnt using the VGG architecture \citep{Simonyan15}.

FCN architecture was considered revolutionary in many aspects. First of all, since FCNs did not include fully connected layers, inference per image was seen to be considerably faster. This was mainly because convolutional layers when compared to fully connected layers, had a marginal number of weights. Second, and maybe more significant, the structure allowed segmentation maps to be generated for images of any resolution. In order to achieve this, FCNs used deconvolutional layers that can upsample coarse deep convolutional layer outputs to dense pixels of any desired resolution. Finally, and most importantly, they proposed the skip architecture for DCNNs.

\begin{figure}[t]
\centering
\includegraphics*[trim=20 0 80 10,clip=false,width=0.98\textwidth]{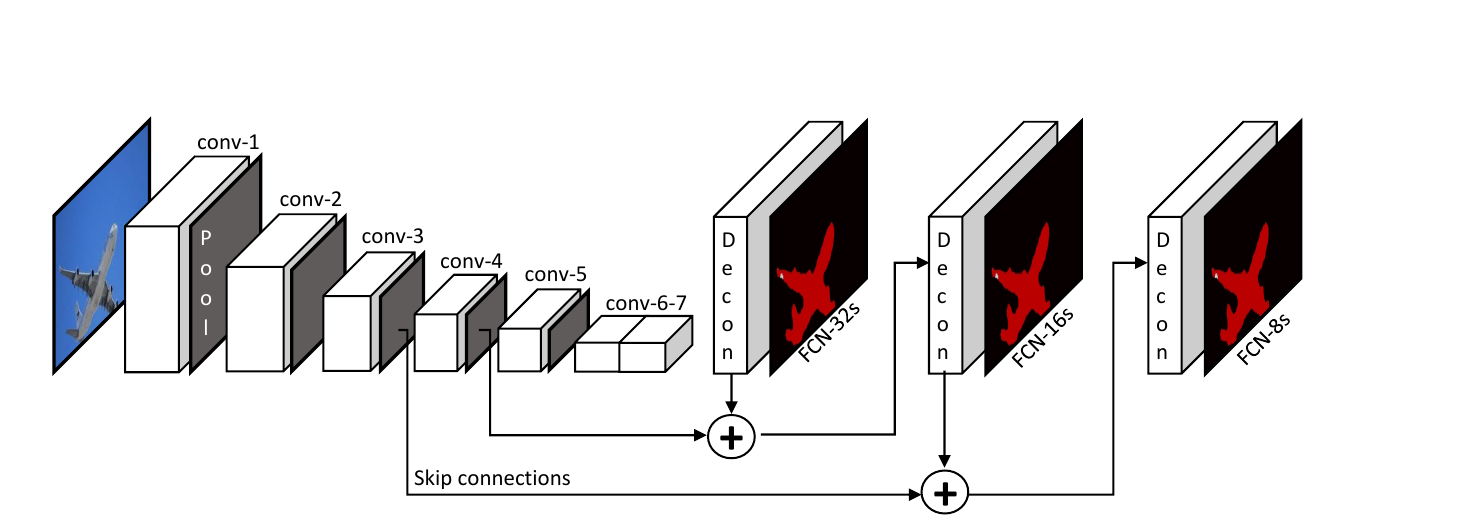}
\caption{Fully convolutional networks (FCNs) are trained end-to-end and are designed to
make dense predictions for per-pixel tasks like semantic segmentation. FCNs consist of no fully connected layers .}
 \label{FCNs}
\end{figure}

Skip architectures (or connections) provide links between nonadjacent layers in DCNNs. Simply by summing or concatenating outputs of unconnected layers, these connections enable information to flow, which would otherwise be lost because of an architectural choice such as max-pooling layers or dropouts. The most common practice is to use skip connections preceding a max-pooling layer, which downsamples layer output by choosing the maximum value in a specific region. Pooling layers helps the architecture create feature hierarchies, but also causes loss of localised information which could be valuable for semantic segmentation, especially at object borders. Skip connections preserve and forward this information to deeper layers by way of bypassing the pooling layers. Actually, the usage of skip connections in \citep{Shelhamer2017}  was perceived as being considerably primitive. The ‘FCN-8s’ and ‘FCN-16s’ networks included these skip connections at different layers. Denser skip connections for the same architecture, namely ‘FCN-4s’ and ‘FCN-2s’, were also utilised for various applications \citep{Zhong2016,Lee2017}. This idea eventually evolved into the encoder-decoder structures \citep{Ronneberger2015,BadrinarayananK15} for semantic segmentation, which are presented in the following section.

\section{Post-FCN Approaches}
Almost all subsequent approaches on semantic segmentation followed the idea of FCNs; thus it would not be wrong to state that decision-making with fully-connected layers effectively ceased to exist\footnote{Many methods utilise fully connected layers such as RCNN \citep{Girshick_2015_ICCV}, which are discussed in the following sections. However, this and other similar methods that include fully connected layers have mostly been succeeded by fully convolutional versions for the sake of computational efficiency.} following the appearance of FCNs to the issue of semantic segmentation.


On the other hand, the idea of FCNs also created new opportunities to further improve deep semantic segmentation architectures. Generally speaking, the main drawbacks of FCNs can be summarised as inefficient loss of label localisation within the feature hierarchy, inability to process global context knowledge, and the lack of a mechanism for multiscale processing. Thus, most subsequent studies have been principally aimed at solving these issues through the proposal of various architectures or techniques. For the remainder of this paper, we analyse these issues under the title, ‘fine-grained localisation’. Consequently, before presenting a list of the post-FCN state-of-the-art methods, we focus on this categorisation of techniques and examine different approaches that aim at solving these main issues. In the following, we also discuss scale invariance in the semantic segmentation context and finish with object detection-based approaches, which are a new breed of solution that aim at resolving the semantic segmentation problem simultaneously with detecting object instances.

\begin{figure}[ht]
\centering
\begin{subfigure}{8cm}	
	\includegraphics*[clip=false,trim=23 20 25 10,width=1\textwidth]{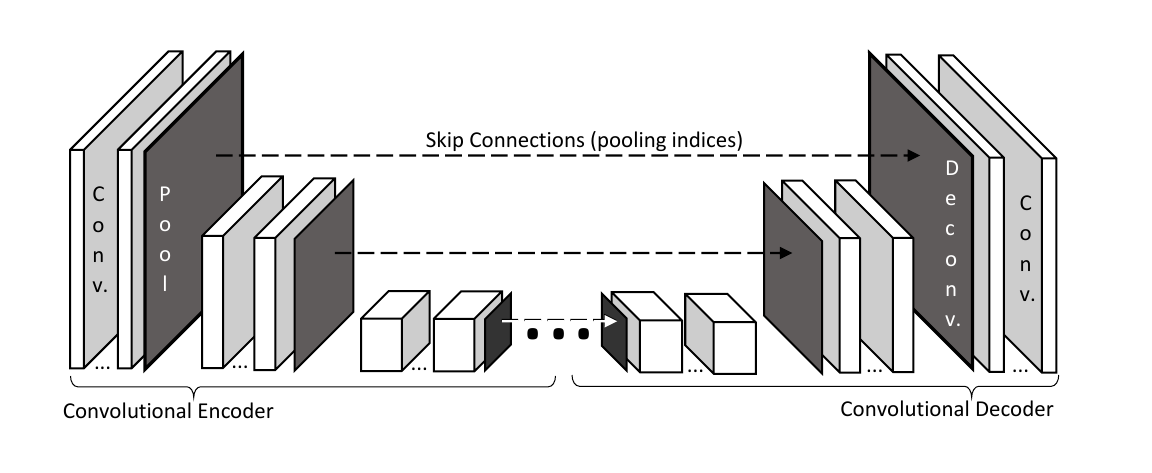}
	\caption{Encoder-Decoder Architecture.}	
\end{subfigure}
\begin{subfigure}{8cm}	
	\includegraphics*[trim=0 0 25 -10,clip=false,width=1\textwidth]{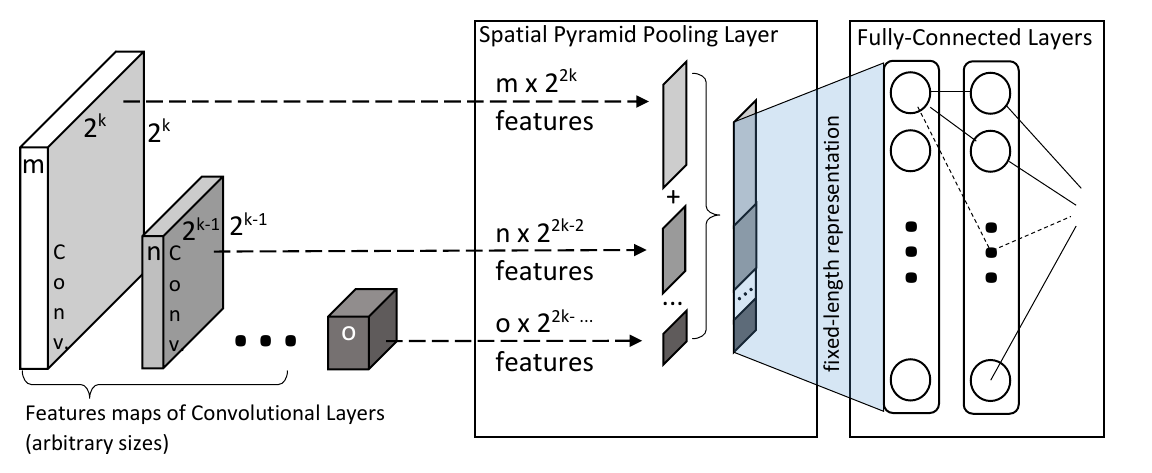}
	\caption{Spatial-Pyramid Pooling Layer}	
\end{subfigure}
\begin{subfigure}{8cm}	
	\includegraphics*[trim=-10 50 0 -30,clip=false,width=1\textwidth]{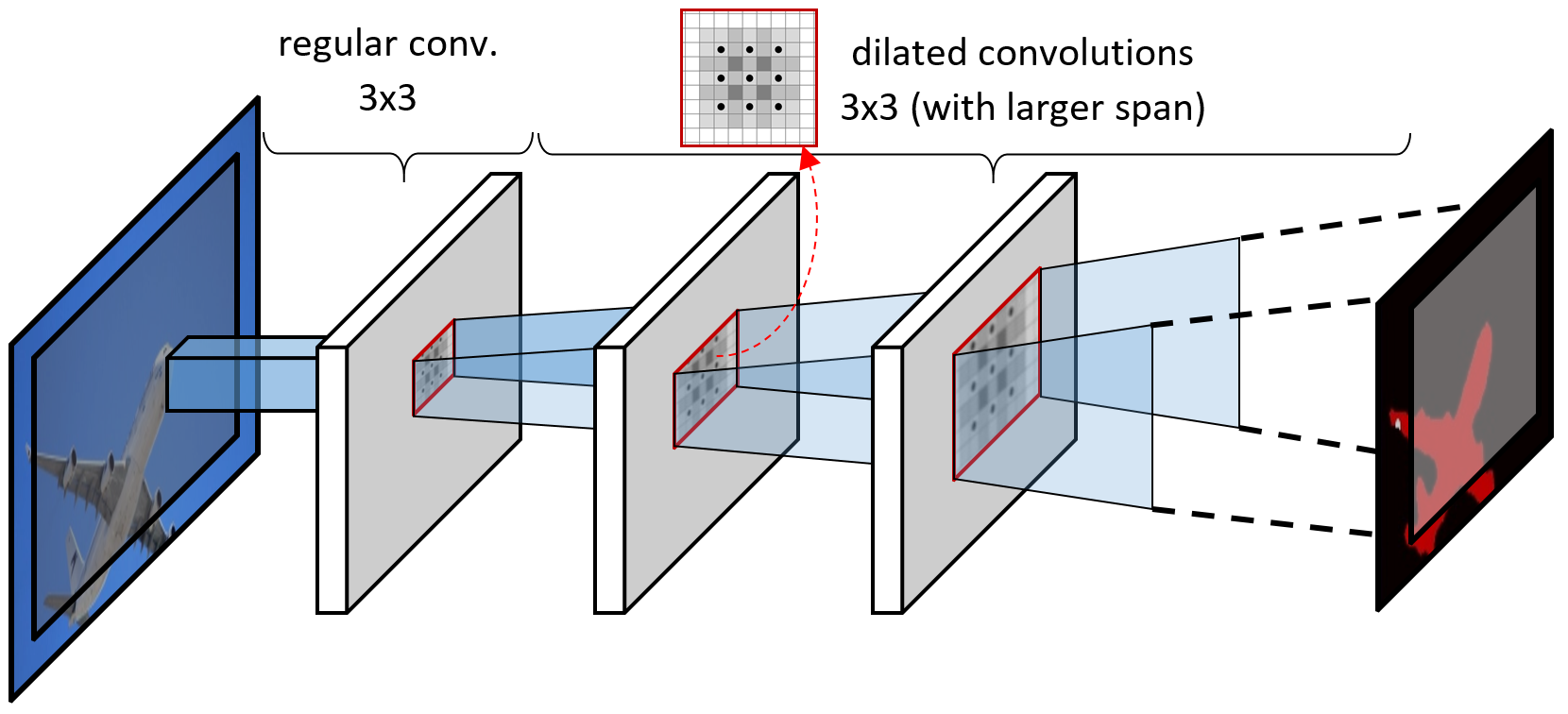}
	\caption{Regular vs. Dilated Convolutions}	
\end{subfigure}
\begin{subfigure}{8cm}	
	\includegraphics*[trim=0 0 0 40,clip=true,width=1\textwidth]{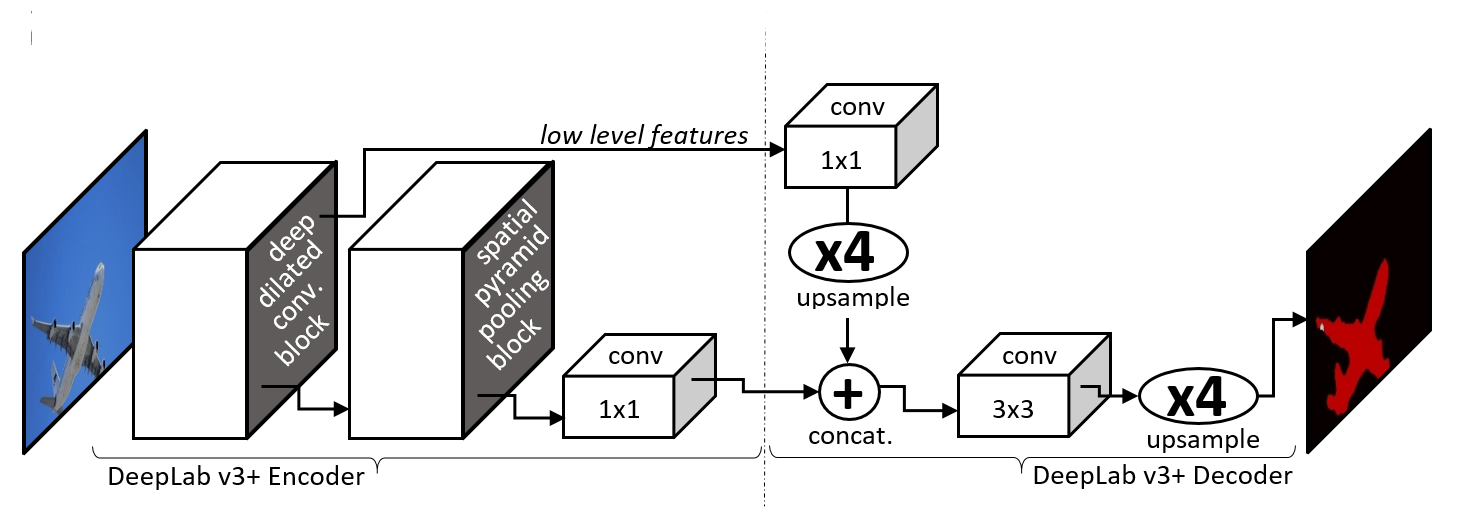}
	\caption{DeepLabv3+ Architecture}	
\end{subfigure}
\begin{subfigure}{8cm}	
	\includegraphics*[trim=0 310 0 0,clip=false,width=1\textwidth]{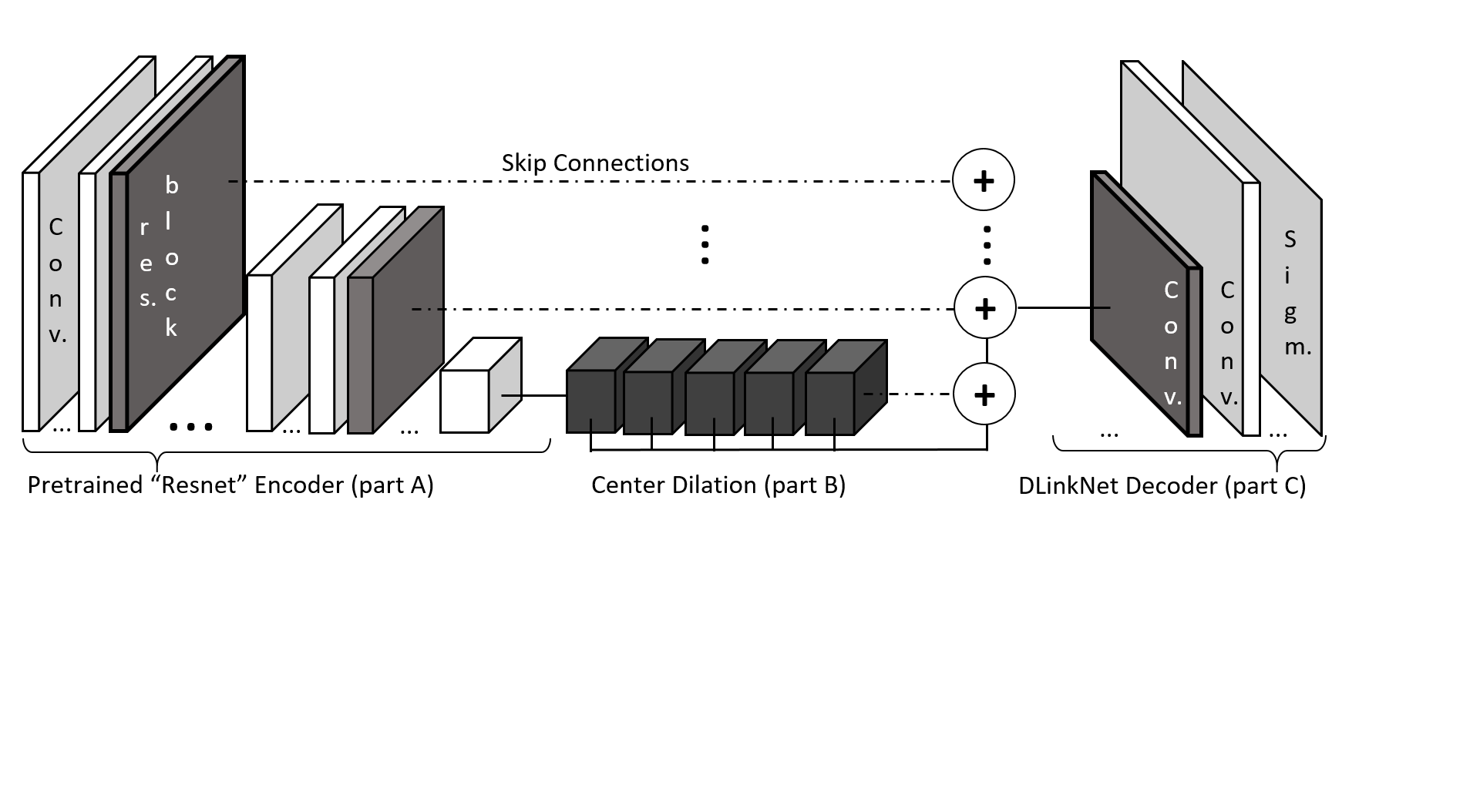}
	\caption{DlinkNet Architecture}	
\end{subfigure}
\caption{Different architectures for fine-grained pixel-wise label localisation.}
\label{Localization}
\end{figure}

\subsection{Techniques for Fine-grained Localisation}
Semantic segmentation is, by definition, a dense procedure; hence it requires fine-grained localisation of class labels at the pixel level. For example, in robotic surgery, pixel errors in semantic segmentation can lead to life or death situations. Hierarchical features created by pooling (i.e., max-pooling) layers can partially lose localisation. Moreover, due to their fully convolutional nature, FCNs do not inherently possess the ability to model global context information in an image, which is also very effective in the localisation of class labels. Thus, these two issues are intertwined in nature, and in the following, we discuss different approaches that aim at overcoming these problems and to providing finer localisation of class labels.

\subsubsection{Encoder-Decoder Architecture}
The so-called Encoder-Decoder (ED) architectures (also known as the U-nets, referring to the pioneering study of \citep{Ronneberger2015}) are comprised of two parts. Encoder gradually reduces the spatial dimension with pooling layers, whilst decoder gradually recovers the object details and spatial dimension. Each feature map of the decoder part only directly receives the information from the feature map at the same level of the encoder part using skip connections, thus EDs can create abstract hierarchical features with fine localisation (see Figure \ref{Localization}.a). U-Net \citep{Ronneberger2015} and Seg-Net \citep{BadrinarayananK15} are very well-known examples. In this architecture, the strongly correlated semantic information, which is provided by the adjacent lower-resolution feature map of the encoder part, has to pass through additional intermediate layers in order to reach the same decoder layer. This usually results in a level of information decay. However, U-Net architectures have proven very useful for the segmentation of different applications, such as medical images \citep{Ronneberger2015}, street view images \citep{BadrinarayananK15}, satellite images \citep{Ulku2019}, just to name a few. Although earlier ED architectures were designed for object segmentation tasks only, there are also modified versions such as ``TernausNetV2'' \citep{Iglovikov18}, that provide instance segmentation capability with minor architectural changes.

\subsubsection{Spatial Pyramid Pooling}
The idea of constructing a fixed-sized spatial pyramid was first proposed by \citep{LazebnikSPP2006}, in order to prevent a Bag-of-Words system losing spatial relations among features. Later, the approach was adopted to CNNs by \citep{KaimingHeSPP}, in that, regardless of the input size, a spatial pyramid representation of deep features could be created in a Spatial Pyramid Pooling Network (SPP-Net). The most important contribution of the SPP-Net was that it allowed inputs of different sizes to be fed into CNNs. Images of different sizes fed into convolutional layers inevitably create different-sized feature maps. However, if a pooling layer, just prior to a decision layer, has stride values proportional to the input size, the feature map created by that layer would be fixed (see Figure \ref{Localization}.b). By \citep{Li18BMVC}, a modified version, namely Pyramid Attention Network (PAN) was additionally proposed. The idea of PAN was combining an SPP layer with global pooling to learn a better feature representation.

There is a common misconception that SPP-Net structure carries an inherent scale-invariance property, which is incorrect. SPP-Net allows the efficient training of images at different scales (or resolutions) by allowing different input sizes to a CNN. However, the trained CNN with SPP is scale-invariant if, and only if, the training set includes images with different scales. This fact is also true for a CNN without SPP layers.

However, similar to the original idea proposed in \citep{LazebnikSPP2006}, the SPP layer in a CNN constructs relations among the features of different hierarchies. Thus, it is quite similar to skip connections in ED structures, which also allow information flow between feature hierarchies.

The most common utilisation of an SPP layer for semantic segmentation is proposed in \citep{KaimingHeSPP}, such that the SPP layer is appended to the last convolutional layer and fed to the pixel-wise classifier.

\subsubsection{Feature Concatenation}
This idea is based on fusing features extracted from different sources. For example, in \citep{Pinheiro2015} the so-called ‘DeepMask’ network utilises skip connections in a feed-forward manner, so that an architecture partially similar to both SPP layer and ED is obtained. The same group extends this idea with a top-down refinement approach of the feed-forward module and propose the so-called ‘SharpMask’ network \citep{Pinheiro2016}, which has proven to be more efficient and accurate in segmentation performance. Another approach from this category is the so-called ‘ParseNet’ \citep{liu2015parsenet}, which fuses CNN features with external global features from previous layers in order to provide context knowledge. Another approach by \citep{Wang2020} is to fuse the ``stage features'' (i.e. deep encoder activations) with ``refinement path features'' (an idea similar to skip connections), using a convolutional (Feature Adaptive Fusion FAF) block.  Although a novel idea in principle, feature fusion approaches (including SPP) create hybrid structures, therefore they are relatively difficult to train. 

\subsubsection{Dilated Convolution} 
The idea of dilated (atrous) convolutions is actually quite simple: with contiguous convolutional filters, an effective receptive field of units can only grow linearly with layers; whereas with dilated convolution, which has gaps in the filter (see Figure \ref{Localization}.c), the effective receptive field would grow much more quickly \citep{Chen18}. Thus, with no pooling or subsampling, a rectangular prism of convolutional layers is created. Dilated convolution is a very effective and powerful method for the detailed preservation of feature map resolutions. The negative aspect of the technique, compared to other techniques, concerns its higher demand for GPU storage and computation, since the feature map resolutions do not shrink within the feature hierarchy \citep{He2016ResNet}.

\subsubsection{Conditional Random Fields} 
As also discussed in Section 3.1.1, CNNs naturally lack mechanisms to specifically ‘focus’ on regions where class intersections occur. Around these regions, graphical models are used to find inference by observing low-level relations between neighbouring feature maps of CNN layers. Consequently, graphical models, mainly CRFs, are utilised as refinement layers in deep semantic segmentation architectures. As in \citep{rother2004}, CRFs connect low-level interactions with output from multiclass interactions, and in this way, global context knowledge is constructed.

As a refinement layer, various methods exist that employ CRFs to DCNNs, such as the Convolutional CRFs \citep{Marvin2018}, the Dense CRF \citep{krahenbuhl2011}, and CRN-as-RNN \citep{ZhengICCV2015}. In general, CRFs help build context knowledge and thus a finer level of localisation in class labels.

\subsubsection{Recurrent Approaches} 
The ability of Recurrent Neural Networks (RNNs) to handle sequential information can help improve segmentation accuracy. For example, \citep{Pfeuffer2019} used Conv-LSTM layers to improve their semantic segmentation results in image sequences. However, there are also methods that use recurrent structures on still images. For example, the Graph LSTM network \citep{Xiaodan2016} is a generalization of LSTM from sequential data or multidimensional data to general graph-structured data for semantic segmentation on 2D still images. Graph-RNN \citep{Shuai2016} is another example of a similar approach in which an LSTM-based network is used to fuse a deep encoder output with the original image in order to obtain a finer pixel-level segmentation. Likewise, in \citep{Lin_2018_ECCV}, the researchers utilised LSTM-chains in order to intertwine multiple scales, resulting in pixel-wise segmentation improvements. There are also hybrid approaches where CNNs and RNNs are fused. A good example of this is the so-called ReSeg model \citep{ReSeg2016}, in which the input image is fed to a VGG-like CNN encoder, and is then processed afterwards by recurrent layers (namely the ReNet architecture) in order to better localise the pixel labels. Another similar approach is the DAG-RNN \citep{Shuai2016}, which utilize a DAG-structured CNN+RNN network, and models long-range semantic dependencies among image units. To the best of our knowledge, no purely recurrent structures for semantic segmentation exist, mainly because semantic segmentation requires a preliminary CNN-based feature encoding scheme.

There is currently an increasing trend in one specific type of RNN, namely ‘recurrent attention modules’. In these modules, attention \citep{Vaswani2017} is technically fused in the RNN, providing a focus on certain regions of the input when predicting a certain part of the output sequence. Consequently, they are also being utilised in semantic segmentation \citep{li2019Emanet,Zhao2018,Oktay2018}.

\begin{figure}[t]
\centering
\begin{subfigure}{12cm}	
	\includegraphics*[trim=0 0 0 0,clip=false,width=1\textwidth]{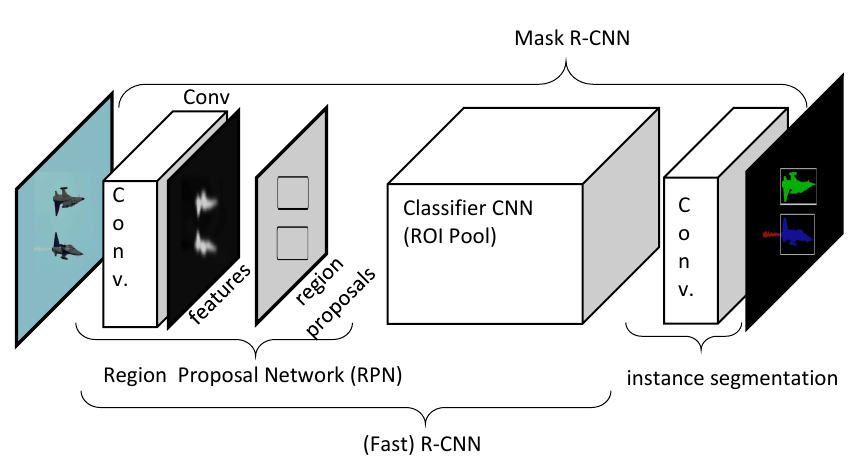}
	\caption{Regions with CNN features-based (Mask-RCNN) architecture}	
\end{subfigure}
\begin{subfigure}{12cm}	
	\includegraphics*[trim=0 0 0 0,clip=false,width=1\textwidth]{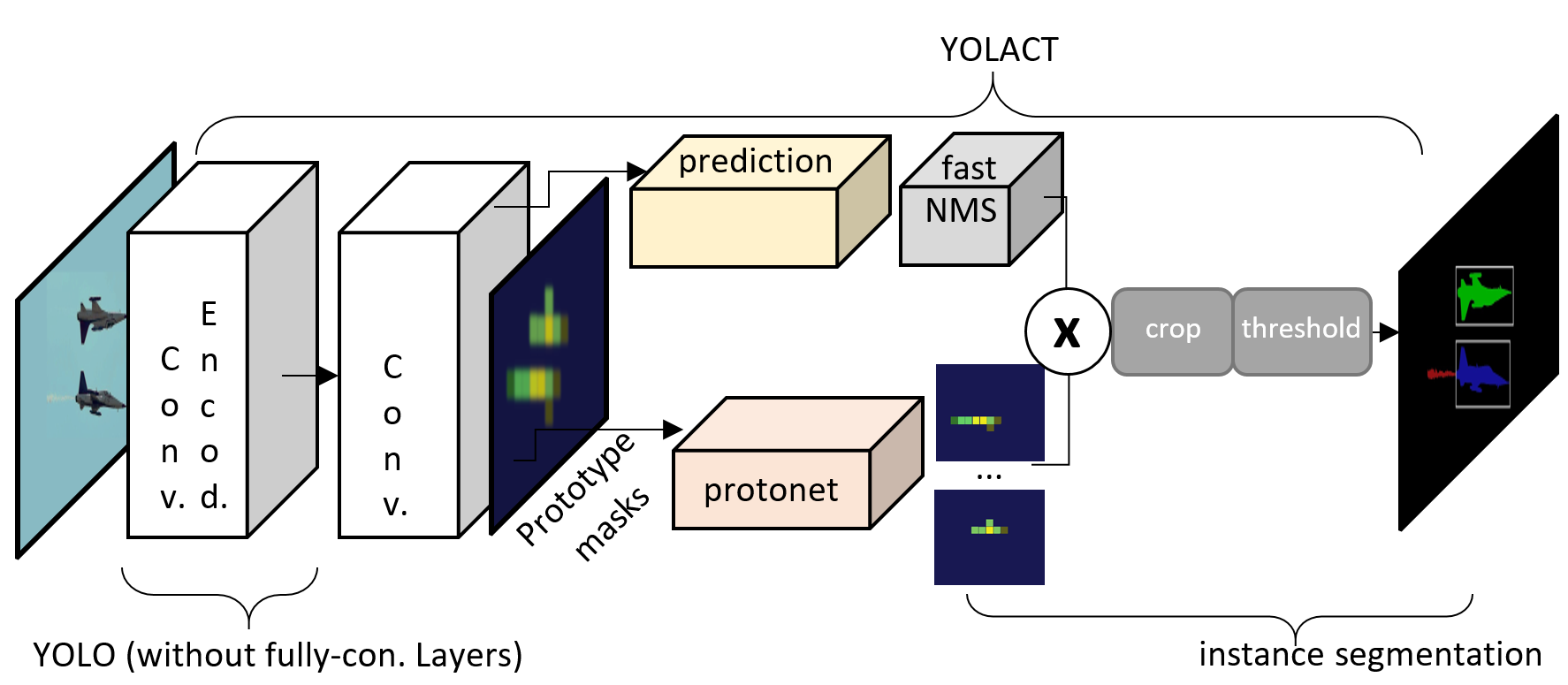}
	\caption{Fully Convolutional Object Detector-based (YOLO)-based architecture}	
\end{subfigure}
\caption{Different architectures for object detection-based semantic segmentation methods}
\label{ObjectBased}
\end{figure}

\subsection{Scale-Invariance}
Scale Invariance is, by definition, the ability of a method to process a given input, independent of the relative scale (i.e. the scale of an object to its scene) or image resolution. Although it is extremely crucial for certain applications, this ability is usually overlooked or is confused with a method’s ability to include multiscale information. A method may use multiscale information to improve its pixel-wise segmentation ability, but can still be dependent on scale or resolution. That is why we find it necessary to discuss this issue under a different title and to provide information on the techniques that provide scale and/or resolution invariance.

In computer vision, any method can become scale invariant if trained with multiple scales of the training set. Some semantic segmentation methods such as \citep{Farabet2013,Eigen2014,Pinheiro2014,Lin2016Efficient,Yu15}  utilise this strategy. However, these methods do not possess an inherent scale-invariance property, which is usually obtained by normalisation with a global scale factor (such as in SIFT by \citep{Lowe2004}). This approach is not usually preferred in the literature on semantic segmentation. The image sets that exist in semantic segmentation literature are extremely large in size. Thus, the methods are trained to memorise that training set, because in principle, overfitting a largescale training set is actually tantamount to solving the entire problem space.

\subsection{Object Detection-based Methods}
There has been a recent growing trend in computer vision, which aims at specifically resolving the problem of object detection, that is, establishing a bounding box around all objects within an image. Given that the image may or may not contain any number of objects, the architectures utilised to tackle such a problem differ to the existing fully-connected/convolutional classification or segmentation models.

\begin{figure*}[t]
\centering
\begin{subfigure}{3.2cm}	
	\includegraphics*[trim=0 0 0 0,clip=false,width=1\textwidth]{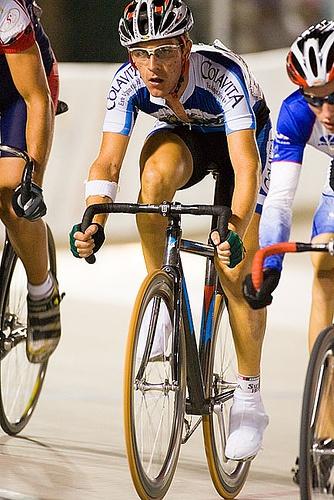}
	\caption{\scriptsize image}	
\end{subfigure}
\begin{subfigure}{3.2cm}	
	\includegraphics*[trim=0 0 0 0,clip=false,width=1\textwidth]{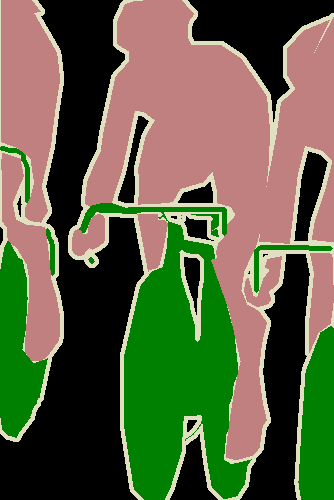}
	\caption{\scriptsize{reference}}	
\end{subfigure}
\begin{subfigure}{3.2cm}	
	\includegraphics*[trim=0 0 0 0,clip=false,width=1\textwidth]{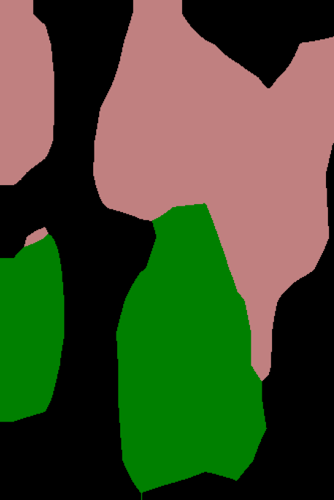}
	\caption{\scriptsize{FCN-32S}}	
\end{subfigure}
\begin{subfigure}{3.2cm}	
	\includegraphics*[trim=0 0 0 0,clip=false,width=1\textwidth]{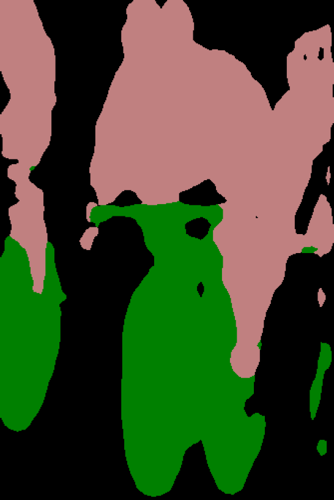}
	\caption{\scriptsize FCN-8S}	
\end{subfigure}
\begin{subfigure}{3.2cm}	
	\includegraphics*[trim=0 0 0 0,clip=false,width=1\textwidth]{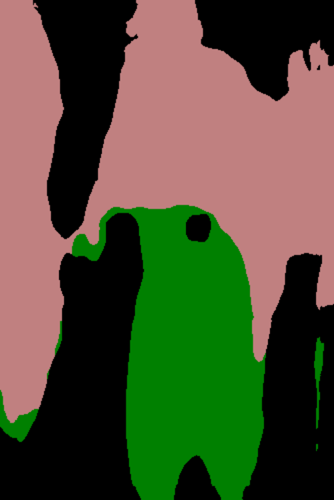}
	\caption{\scriptsize CMSA}	
\end{subfigure}
\begin{subfigure}{3.2cm}	
	\includegraphics*[trim=0 0 0 0,clip=false,width=1\textwidth]{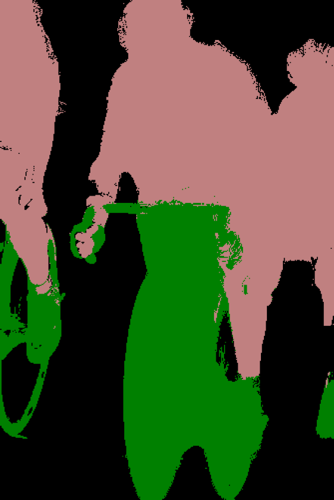}
	\caption{\scriptsize{DeepLabv1}}	
\end{subfigure}
\begin{subfigure}{3.2cm}	
	\includegraphics*[trim=0 0 0 0,clip=false,width=1\textwidth]{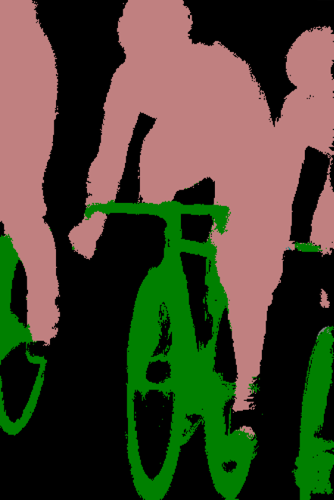}
	\caption{\scriptsize{CRF-as-RNN}}	
\end{subfigure}
\begin{subfigure}{3.2cm}	
	\includegraphics*[trim=0 0 0 0,clip=false,width=1\textwidth]{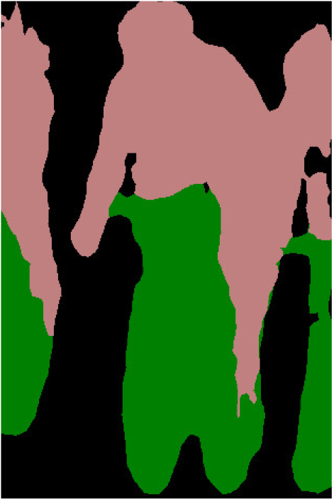}
	\caption{\scriptsize{DeepLab.v2}}	
\end{subfigure}
\begin{subfigure}{3.2cm}	
	\includegraphics*[trim=0 0 0 0,clip=false,width=1\textwidth]{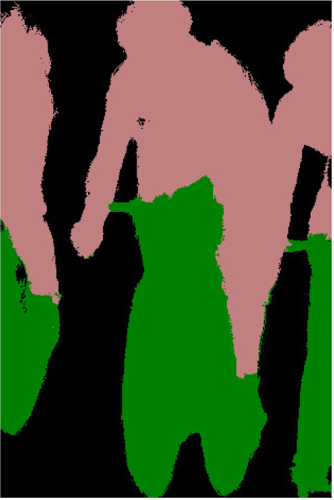}
	\caption{\scriptsize{DeepLab.v2+CRF}}	
\end{subfigure}
\begin{subfigure}{3.2cm}	
	\includegraphics*[trim=0 0 0 0,clip=false,width=1\textwidth]{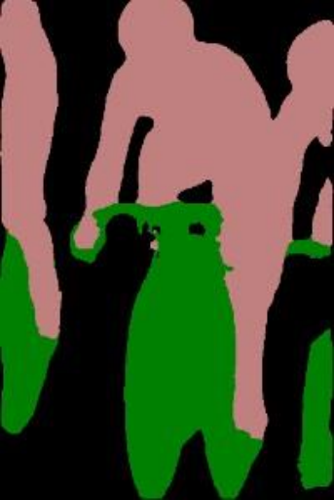}
	\caption{\scriptsize{PAN}}	
\end{subfigure}
\caption{(a) A sample image from the PASCAL VOC validation set, (b) its semantic segmentation ground truth, and results obtained from different studies are depicted: c) FCN-32S \citep{Shelhamer2017}, d) FCN-8S \citep{Shelhamer2017}, e) CMSA \citep{Eigen2014}, f) DeepLab-v1 \citep{Chen14}, g) CRF-as-RNN \citep{ZhengICCV2015}, h) DeepLab-v2 \citep{Chen18}, i) DeepLab-v2 with CRF refinement \citep{Chen18}, j) PAN \citep{Li18BMVC}.}
\label{SampleResults}
\end{figure*}

The pioneering study that represents this idea is the renowned ‘Regions with CNN features’ (RCNN) network \citep{RCNN2014}. Standard CNNs with fully convolutional and fully connected layers lack the ability to provide varying length output, which is a major flaw for an object detection algorithm that aims to detect an unknown number of objects within an image. The simplest way to resolve this problem is to take different regions of interest from the image, and then to employ a CNN in order to detect objects within each region separately. This region selection architecture is called the ‘Region Proposal Network’ (RPN) and is the fundamental structure used to construct the RCNN network (see Figure \ref{ObjectBased}.a). Improved versions of RCNN, namely ‘Fast-RCNN’ \citep{RCNN2014} and ‘Faster-RCNN’ \citep{NIPS2015_FasterRCNN} were subsequently also proposed by the same research group. Because these networks allow for the separate detection of all objects within the image, the idea was easily implemented for instance segmentation, as the ‘Mask-RCNN’ \citep{He2017MaskR}.

The basic structure of RCNNs included the RPN, which is the combination of CNN layers and a fully connected structure in order to decide the object categories and bounding box positions. As discussed within the previous sections of this paper, due to their cumbersome structure, fully connected layers were largely abandoned with FCNs. RCNNs shared a similar fate when the ‘You-Only-Look-Once’ (YOLO) by \citep{YOLO2016} and ‘Single Shot Detector’ (SSD) by \citep{SSD16} were proposed. YOLO utilises a single convolutional network that predicts the bounding boxes and the class probabilities for these boxes. It consists of no fully connected layers, and consequently provides real-time performance. SSD proposed a similar idea, in which bounding boxes were predicted after multiple convolutional layers. Since each convolutional layer operates at a different scale, the architecture is able to detect objects of various scales. Whilst slower than YOLO, it is still considered to be faster then RCNNs. This new breed of object detection techniques was immediately applied to semantic segmentation. Similar to MaskRCNN, ‘Mask-YOLO’ \citep{maskyolo2019} and ‘YOLACT’ \citep{YOLACT2019} architectures were implementations of these object detectors to the problem of instance segmentation (see Figure \ref{ObjectBased}b). Similar to YOLACT, some other methods also achieve fast, real-time instance segmentation such as; ESE-Seg \citep{xu2019explicit}, SOLO\citep{wang2020solo}, SOLOv2\citep{wang2020solov2}, DeepSnake \citep{peng2020deep}, and CenterPoly\citep{perreault2021centerpoly}.

Locating objects within an image prior to segmenting them at the pixel level is both intuitive and natural, due to the fact that it is effectively how the human brain supposedly accomplishes this task \citep{Rosenholtz2016}.  In addition to these ``two-stage (detection+segmentation) methods, there are some recent studies that aim at utilizing the segmentation task to be incorporated into one-stage bounding-box detectors and result in a simple yet efficient instance segmentation framework \citep{Xu19ICCV,Zhang2020, Lee2020,Xie2020}. However, the latest trend is to use global-area-based methods by generating intermediate FCN feature maps and then assembling these basis features to obtain final masks \citep{chen2020blendmask, kim2021devil, ke2021deep}.  

In recent years, a trend of alleviating the demand for pixel-wise labels is realized mainly by employing bounding boxes, and by expanding from semantic segmentation to instance segmentation applications. In both semantic segmentation and instance segmentation methods, the category of each pixel is recognized, and the only difference is that instance segmentation also differentiates object occurrences of the same category. Therefore, weakly-supervised instance segmentation (WSIS) methods are also utilized for instance segmentation. The supervision of WSIS methods can use different annotation types for training, which are usually in the form of either bounding boxes \citep{khoreva2017simple, hsu2019weakly, arun2020weakly, tian2021boxinst, lee2021bbam, cheng2021per} or image-level labels \citep{ liu2020leveraging, shen2021parallel, zhou2016learning, shen2021toward}. Hence, employing object detection-based methods for semantic segmentation is an area significantly prone to further development in near future by the time this manuscript is prepared. 

\begin{table*}[p]
\centering
\begin{tabular}{|p{2.5cm}|p{9.3cm}|p{2.9cm}|>{\centering\arraybackslash}p{0.8cm}|}
\hline
Method & Method Summary & Rankings & Eff.\\\hline \hline
MultiScale-Net.\newline\scriptsize\citep{Farabet2013} & \emph{Multiscale convolutional network fused parallel with a segmentation framework (either superpixel or CRF-based). Relatively lower computational efficiency due to a CRF block.} & \scriptsize68.7\% mPA @SIFTflow & $\star$$\hspace{0.1cm}$$\star$\\\hline
Recurrent CNN\newline\scriptsize\citep{Pinheiro2014} & \emph{Recurrent architecture constructed by using different instances of a CNN, in which each network instance is fed with previous label predictions (obtained from the previous instance). Heavy computational load when multiple instances (3 in their best performing experiments) are fed.}  & \scriptsize77.7\% mPA @SIFTflow  & $\star$ \\\hline
FCN\newline\citep{Shelhamer2017} \newline  &  \emph{Fully convolutional encoder structure (i.e., no fully connected layers) with skip connections that fuse multiscale activations at the final decision layer. Relative fast due to no fully connected layers or a refinement block.} & \scriptsize85.2\% mPA @SIFTflow\newline\scriptsize62.2\% mIoU @PASCAL 2012\newline65.3\% mIoU @CitySca. (w/o course)\newline39.3\% mIoU @ADE20K   & $\star$ $\star$ $\star$ \\\hline
DeepLab.v1\newline\citep{Chen14}  & \emph{CNN with dilated convolutions, succeeded by a fully-connected (i.e. Dense) CRF.} Fast and optimized computation leads to near real-time performance. & \scriptsize66.4\% mIoU @PASCAL 2012 & $\star$ $\star$ $\star$\\\hline
CMSA\newline\citep{Eigen2014} \newline  & \emph{Layers of a pyramidal input are fed to separate FCNs for different scales in parallel. These multiscale FCNs are also connected in series to provide pixel-wise category, depth and normal output, simultaneously. Relatively lower computational efficiency due to progressive processing of sequence of different scales}. & \scriptsize83.8\% mPA @SIFTflow\newline\scriptsize62.6\% mIoU @PASCAL 2012 & $\star$$\hspace{0.1cm}$$\star$ \\\hline
UNet\newline\citep{Ronneberger2015}  & \emph{Encoder/decoder structure with skip connections that connect same levels of ED and final input-sized classification layer. Efficient computation load due to no fully connected layers or a refinement block.} & \scriptsize72.7\% mIoU @PASCAL 2012 (\emph{tested by \citep{Zhang_2018})}& $\star$ $\star$ $\star$\\\hline
SegNet\newline\citep{BadrinarayananK15}\newline  & \emph{Encoder/decoder structure (similar to UNet) with skip connections that transmit only pooling indices (unlike U-Net, for which skip connections concatenate same-level activations.  Efficient computation load due to no fully connected layers or a refinement block).} & \scriptsize59.9\% mIoU @PASCAL 2012\newline79.2\% mIoU @CitySca. (w/o course) & $\star$ $\star$ $\star$\\\hline
DeconvNet\newline\citep{Noh2015}\newline  & \emph{Encoder/decoder structure (namely ‘the Conv./Deconv. Network’) without skip connections. The encoder (convolutional) part of the network is transferred from the VGG-VD-16L. Efficient computation load due to no fully connected layers or a refinement block.} \citep{Simonyan15}. & \scriptsize74.8\% mIoU @PASCAL 2012 & $\star$ $\star$ $\star$ \\\hline
MSCG \newline\citep{Yu15} \newline& \emph{Multiscale context aggregation using only a rectangular prism of dilated convolutional layers, without pooling or subsampling layers, to perform pixel-wise labelling. Efficient computation load due to no fully connected layers or a refinement block.} & \scriptsize67.6\% mIoU @PASCAL 2012\newline\scriptsize67.1\% mIoU @CitySca. (w/o course) &  $\star$ $\star$ $\star$ \\\hline
CRF-as-RNN\newline\citep{ZhengICCV2015}  \newline  & \emph{Fully convolutional CNN (i.e., FCN) followed by a CRF-as-RNN layer, in which an iterative CRF algorithm is formulated as an RNN. Because of the RNN block, computational efficiency is limited.}   & \scriptsize65.2\% mIoU @PASCAL 2012\newline\scriptsize62.5\% mIoU @CitySca. (w/o course) &  $\star$$\hspace{0.1cm}$$\star$ \\\hline
FeatMap-Net.\newline\citep{Lin2016Efficient} \newline  & \emph{Layers of a pyramidal input fed to parallel multiscale feature maps (i.e., CNNS), and later fused in an upsample/concatenation (i.e. pyramid pooling) layer  to provide the final feature map to a Dense CRF Layer. Well-planned but loaded architecture leads to moderate computational efficiency.}  & \scriptsize88.1\% mPA @SIFTflow\newline\scriptsize75.3\% mIoU @PASCAL 2012 & $\star$$\hspace{0.1cm}$$\star$\\\hline 
Graph LSTM\newline\citep{Xiaodan2016} \newline  & \emph{Generalization of LSTM from sequential data to general graph-structured data for semantic segmentation on 2D still images, mostly people/parts. Graph-LSTM processing considerably limits computation efficiency.} & \scriptsize60.2\% mIoU @PASCAL Person/Parts 2010 & $\star$\\\hline
DAG-RNN\newline\citep{Shuai2016} \newline  & \emph{DAG-structured CNN+RNN network that models long-range semantic dependencies among image units. Due to chain structured sequential processing of pixels with a recurrent model, the computational efficiency is considerably limited.} & \scriptsize85.3\% mPA @SIFTflow & $\star$\\\hline
\end{tabular}
\end{table*}

\begin{table*}[p]
\centering
\begin{tabular}{|p{2.5cm}|p{9.3cm}|p{2.9cm}|>{\centering\arraybackslash}p{0.8cm}|}
\hline
cont'd. &  &  & \\\hline \hline
DeepLab.v2\newline\citep{Chen18}   \newline  & \emph{Improved version of DeepLab.v1, with additional ‘dilated (atrous) spatial pyramid pooling’ (ASPP) layer. Similar computational performance to DeepLab.v1.} & \scriptsize79.7\% mIoU @PASCAL 2012\newline\scriptsize70.4\% mIoU @CitySca. (w/o course) & $\star$ $\star$ $\star$ \\\hline
PSPNet\newline\citep{ZhaoCVPR2017} \newline   & \emph{CNN followed by a pyramid pooling layer similar to \citep{KaimingHeSPP}, but without a fully connected decision layer. Hence, computational performance closer to FCN \citep{Shelhamer2017}.} & \scriptsize85.5\% mIoU @PASCAL 2012\newline81.2\% mIoU @CitySca. (w. course)\newline55.4\% mIoU @ADE20K & $\star$ $\star$ $\star$ \\\hline
DeepLab.v3\newline\citep{Chen2017} & \emph{Improved version of DeepLab.v2, with optimisation of ASPP layer hyperparameters and without a Dense CRF layer, for faster operation.}  & \scriptsize85.7\% mIoU @PASCAL 2012\newline81.3\% mIoU @CitySca. (w. course) & $\star$ $\star$ $\star$ \\\hline
DIS\newline\citep{Luo2017ICCV} \newline  & \emph{One network predicts labelmaps/tags, while another performs semantic segmentation using these predictions. Both networks use ResNet101 \citep{He2016ResNet} for preliminary feature extraction. They declare similar computational efficiency to DeepLabv2 \citep{Chen18}} &  \scriptsize41.7\% mIoU @COCO\newline\scriptsize86.8\% mIoU @PASCAL 2012&  $\star$ $\star$ $\star$ \\\hline
Mask-RCNN\newline\citep{He2017MaskR} \newline   & \emph{Object Detector Fast-RCNN followed by ROI-pooling and Convolutional layers, applied to instance segmentation, with near real-time performance (see Figure \ref{ObjectBased}.a).}  & \scriptsize37.1\% mIoU @COCO \emph{tested by} \citep{YOLACT2019} & $\star$ $\star$ $\star$ \\\hline
GCN\newline\citep{Peng2017GCN} \newline   & \emph{Fed by an initial ResNet-based \citep{He2016ResNet} encoder, GCN uses large kernels to fuse high- and low-level features in a multiscale manner, followed by a convolutional Border Refinement (BR) module. Its fully convolutional architectue allows near real-time performance. }  & \scriptsize83.6\% mIoU @PASCAL 2012\newline76.9\% mIoU @CitySca. (w/o course) & $\star$ $\star$ $\star$\\\hline
SDN\newline\citep{Fu_2019}  \newline  & \emph{UNET architecture that consists of multiple shallow deconvolutional networks, called SDN units, stacked one by one to integrate contextual information and guarantee fine recovery of localised information. Computational efficiency similar to UNET-like architectures.} & \scriptsize83.5\% mIoU @PASCAL 2012 & $\star$ $\star$ $\star$ \\\hline
DFN\newline\citep{Yu2018CVPR} \newline  &  \emph{Consists of two sub-networks: Smooth Net (SN) and Border Net (BN). SN utilises an attention module and handles global context, whereas BN employs a refinement block to handle borders. Limited computational efficiency due to an attention block}. & \scriptsize86.2\% mIoU @PASCAL 2012\newline\scriptsize80.3\% mIoU @CitySca. (w.course) & $\star$$\hspace{0.1cm}$$\star$ \\\hline
MSCI\newline\citep{Lin_2018_ECCV} \newline   &\emph{Aggregates features from different scales via connections between Long Short-term Memory (LSTM) chains. Limited computational efficiency due to multiple RNN blocks (i.e. LSTMs).}  & \scriptsize88.0\% mIoU @PASCAL 2012 & $\star$$\hspace{0.1cm}$$\star$ \\\hline
DeepLab.v3+\newline\citep{Liang2018a}  \newline   &  \emph{Improved version of DeepLab.v3, using special encoder-decoder structure with dilated convolutions (with no Dense CRF employed for faster operation).} & \scriptsize87.3\% mIoU @PASCAL 2012\newline82.1\% mIoU @CitySca. (w. course) & $\star$ $\star$ $\star$\\\hline
HPN\newline\citep{Shi2018}  \newline  & \emph{Followed by a convolutional ‘Appearance Feature Encoder’, a ‘Contextual Feature Encoder’ consisting of LSTMs generates super-pixel features fed to a Softmax-based classification layer. Limited computational efficiency due to multiple LSTMs.} & \scriptsize85.8\% mIoU @PASCAL 2012\newline\scriptsize92.3\% mPA @SIFTflow & $\star$$\hspace{0.1cm}$$\star$ \\\hline
EncNet\newline\citep{Zhang_2018_CVPR}   \newline  & \emph{Fully connected structure to extract context is fed by dense feature maps (obtained from ResNet \citep{He2016ResNet}) and followed by a convolutional prediction layer. Fully connected layers within their ``Context Encoding Module'' limits computational performance.}  & \scriptsize85.9\% mIoU @PASCAL 2012\newline55.7\% mIoU @ADE20K & $\star$$\hspace{0.1cm}$$\star$ \\\hline
PSANet\newline\citep{Zhao2018} &  \emph{A convolutional point-wise spatial attention (PSA) module is attached to o pretrained convolutional encoder, so that pixels are interconnected through a self-adaptively learnt attention map to provide global context. Additional PSA module limits computational efficieny compared to fully convolutional architectures (e.g. FCN).} & \scriptsize85.7\% mIoU @PASCAL 2012\newline81.4\% mIoU @CitySca. (w. course) & $\star$$\hspace{0.1cm}$$\star$ \\\hline
\end{tabular}
\end{table*}

\begin{table*}[p]
\centering
\begin{tabular}{|p{2.5cm}|p{9.3cm}|p{2.9cm}|p{0.8cm}|}
\hline
cont'd. &  &  & \\\hline \hline
PAN\newline\citep{Li18BMVC} &  \emph{SPP layer with global pooling architecture. Similar architecture and thus, computational efficiency with PSPNet \citep{ZhaoCVPR2017}.} & \scriptsize84.0\% mIoU @PASCAL 2012 (\emph{taken from the paper, not listed in the leaderboard}) & $\star$ $\star$ $\star$ \\\hline
ExFuse\newline\citep{Zhang_2018}  \newline   &  \emph{Improved version of GCN \citep{Peng2017GCN} for feature fusing which introduces more semantic information into low-level features and more spatial details into high-level features, by additional skip connections. Computational performance comparable to GCN.} & \scriptsize87.9\% mIoU @PASCAL 2012 & $\star$ $\star$ $\star$ \\\hline
EMANet152\newline\citep{li2019Emanet}  \newline  &  \emph{Novel attention module between two CNN structures converts input feature maps to output feature maps, thus providing global context. Computationally more efficient compared to other attention governing architectures (e.g. PSANet).} & \scriptsize88.2\% mIoU @PASCAL 2012\newline39.9\% mIoU @COCO & $\star$ $\star$ $\star$ \\\hline
KSAC\newline\citep{huang2019}  \newline  &  \emph{Allows branches of different receptive fields to share the same kernel to facilitate communication among branches and perform feature augmentation inside the network. The idea is similar to ASPP layer of DeepLabv3 \citep{Chen2017}, hence similar computational performance.} &  \scriptsize88.1\% mIoU @PASCAL 2012 & $\star$ $\star$ $\star$ \\\hline
CFNet\newline\citep{Zhang_2019_CVPR}  \newline  & \emph{Using a distribution of co-occurrent features for a given target in an image, a fine-grained spatial invariant representation is learnt and the CFNet is constructed. Similar architecture to PSANet \citep{Zhao2018}, hence similar (and limited) computational performance due to fully connected layers.} & \scriptsize87.2\% mIoU @PASCAL 2012 & $\star$$\hspace{0.1cm}$$\star$ \\\hline
YOLACT\newline\citep{YOLACT2019}  & \emph{Object Detector YOLO followed by Class Probability and Convolutional layers, applied to instance segmentation (see Figure \ref{ObjectBased}.b), with \underline{real-time} semantic segmentation performance}.  & \scriptsize72.3\% mAP$_{50}$ @PASCAL SBD\newline\scriptsize31.2\% mAP @COCO & $\star$ $\star$ $\star$\newline$\star$ \\\hline
ESE-Seg\newline\citep{Xu19ICCV}  & \emph{ESE-Seg is an object detection-based approach that uses explicit shape encoding by explicitly decoding the multiple object shapes with tensor operations in real-time.}  & \scriptsize69.3\% mAP$_{50}$ @PASCAL SBD\newline\scriptsize21.6\% mAP @COCO & $\star$ $\star$ $\star$\newline$\star$ \\\hline
SOLO\newline\citep{wang2020solo}  \newline  & \emph{The central idea of SOLO framework is to reformulate the instance segmentation as two simultaneous problems: category prediction and instance mask generation, using a single convolutional backbone. The model can run in real-time with proper parameter tuning.} & \scriptsize37.8\% mAP @COCO & $\star$ $\star$ $\star$\\\hline
EfficientNet-L2 + NASFPN + Noisy Student\newline\citep{ZophGLCLC020}  & \emph{The study aims at understaing the effect of pre- and self training and apply this to semantic segmentation problem. For their experiment, they utilize a neural architecture search (NAS) strategy \citep{Ghiasi2019}
with EfficientNet-L2 \citep{Xie2020cvpr} as the backbone architecture. The model is the leader of PASCAL VOC 2012 challenge by the time this manuscript was written.} & \scriptsize90.5\% mIoU @PASCAL 2012 &  $\star$ $\star$ $\star$ \\\hline
DCNAS\newline\citep{zhang2020dcnas}  \newline  & \emph{Neural Architecture Search applied to MobileNetV3 \citep{Howard2019}, a densely connected search space for semantic segmentation. Although computational performance is not explicitly indicated, the resulting architecture possibly provides U-Net like computational efficiency for model inference. } & \scriptsize86.9\% mIoU @PASCAL 2012 (\emph{taken from the paper, not listed in the leaderboard})\newline83.6\% mIoU @CitySca. (w. course)& $\star$ $\star$ $\star$\\\hline
SOLOv2\newline\citep{wang2020solov2}  \newline  & \emph{Updated, real-time version of SOLO \citep{wang2020solo}, empowered by an efficient and holistic instance mask representation scheme, which dynamically segments each instance in the image, without resorting to bounding
box detection.} & \scriptsize37.1\% mAP @COCO& $\star$ $\star$ $\star$\newline$\star$\\\hline
\end{tabular}
\label{MethodsTable}
\end{table*}

\begin{table*}[h]
\centering
\begin{tabular}{|p{2.5cm}|p{9.3cm}|p{2.9cm}|p{0.8cm}|}
\hline
cont'd. &  &  & \\\hline \hline
Deep Snake\newline\citep{peng2020deep}  \newline  & \emph{Deep Snake is a fully convolutional architecture with a contour-based approach for real-time instance segmentation.} & \scriptsize62.1\% mAP$_{50}$ @PASCAL SBD\newline\scriptsize30.3\% mAP @COCO & $\star$ $\star$ $\star$\newline$\star$\\\hline
BlendMask\newline\citep{chen2020blendmask}  \newline  & \emph{Using both top-down and
bottom-up instance segmentation approaches, BlendMask learns attention maps for each instance using a single
convolution layer.} & \scriptsize37.1\% mAP @COCO & $\star$ $\star$ $\star$\newline$\star$\\\hline
SwiftNetRN18-Pyr\newline\citep{orvsic2021efficient}  \newline  & \emph{Based on shared pyramidal representation and fusion of heterogeneous features, SwiftNetRN18-Pry fuses hybrid representation within a ladder-style decoder. Provides beyond real-time performance with modest accuracy.} & \scriptsize35.0\% mIoU @ADE20K & $\star$ $\star$ $\star$\newline$\star$\\\hline
BOXInst\newline\citep{tian2021boxinst}  \newline  & \emph{Achieves mask-level instance segmentation with only bounding-box
annotations for training. Core idea is to redesign the loss of learning masks in instance segmentation} & \scriptsize61.4\% mAP$_{50}$ @PASCAL SBD\newline\scriptsize31.6\% mAP @COCO & $\star$ $\star$ $\star$\\\hline
\end{tabular}
\caption{State-of-the-art semantic segmentation methods, showing the method name and reference, brief summary, problem type targeted, and refinement model (if any).}
\label{MethodsTable}
\end{table*}

\subsection{Evolution of Methods}
In Table \ref{MethodsTable}, we present several semantic segmentation methods, each with a brief summary, explaining the fundamental idea that represents the proposed solutions, their position in available leaderboards, and a categorical level of the method's computational efficiency. The intention is for readers to gain a better evolutionary understanding of the methods and architectures in this field, and a clearer conception of how the field may subsequently progress in the future. Regarding the brief summaries of the listed methods, please refer to the categorisations provided earlier in this section.

Table \ref{MethodsTable} includes 34 methods spanning an eight-year period, starting with early deep learning approaches through to the most recent state-of-the-art techniques. Most of the listed studies have been quite successful and have significantly high rankings in the previously mentioned leaderboards. Whilst there are many other methods, we believe this list to be a clear depiction of the advances in deep learning-based semantic segmentation approaches. In Figure \ref{SampleResults}, a sample image from the PASCAL VOC validation set, its semantic segmentation ground truth and results obtained from some of the listed studies are depicted. Figure \ref{SampleResults} clearly shows the gradually growing success of different methods starting with the pioneering FCN architectures to more advanced architectures such as DeepLab \citep{Chen14,Chen18} or CRF-as-RNN \citep{ZhengICCV2015}.

Judging by the picture it portrays, the deep evolution of the literature clearly reveals a number of important implications. First, graphical model-based refinement modules are being abandoned due to their slow nature. A good example of this trend would be the evolution of DeepLab from \citep{Chen14} to \citep{Liang2018a} (see Table \ref{MethodsTable}). Notably, no significant study published in 2019 and 2020 employed a CRF-based or similar module to refine their segmentation results. Second, most studies published in the past two years show no significant leap in performance rates. For this reason, researchers have tended to focus on experimental solutions such as object detection-based or Neural Architecture Search (NAS)-based approaches. Some of these very recent group of studies \citep{zhang2020dcnas,ZophGLCLC020} focus on (NAS)-based techniques, instead of hand-crafted architectures. EfficientNet-NAS \citep{ZophGLCLC020} belongs to this category and is the leading study in PASCAL VOC 2012 semantic segmentation challenge at the time the paper was prepared. We believe that the field will witness an increasing interest in NAS-based methods in the near future. In general, considering all studies of the post-FCN era, the main challenge of the field still remains to be \emph{efficiently} integrating (i.e. in  real-time) global context to localisation information, which still does not appear to have an off-the-shelf solution, although there are some promising techniques, such as YOLACT \citep{YOLACT2019}.

In Table \ref{MethodsTable}, the right-most column represents a categorical level of computational efficiency. 
We use a four-level categorisation (one star to four stars) to indicate the computational efficiency of each listed method. For any assigned level of the computational efficiency of a method, we explain our reasoning in the table with solid arguments. For example, one of the four-star methods in Table \ref{MethodsTable} is ``YOLACT'' by \citep{YOLACT2019}, which claims to provide real-time performance (i.e. $>$30fps) on both PASCAL VOC 2012 and COCO image sets. 

\section{Future Scope and Potential Research Directions}
Although tremendous successes have been achieved so far in the semantic segmentation field, there are still many open challenges in this field due to hard requirements time-consuming pixel-level annotations, lack of generalization ability to new domains and classes, and need for real-time performance with higher segmentation accuracies. In this section, we categorize possible future directions under different titles by providing examples of recent studies that represent that direction.

\subsection{Weakly-Supervised Semantic Segmentation (WSSS)}
Over the last few years, there has been an increasing research effort directed towards the approaches that are alternative to pixel-level annotations such as; unsupervised, semi-supervised \citep{he2021re} and weakly-supervised methods. Recent studies show that, WSSS methods usually perform better than the other schemes \citep{chan2021comprehensive} where annotations are in the form of image-level labels \citep{kolesnikov2016seed, pathak2015constrained, pinheiro2015image, wang2020self, ahn2018learning, li2021pseudo, chang2020weakly, xu2021leveraging, yao2021non, jiang2021online}, video-level labels \citep{ zhong2016weakly}, scribbles \citep{lin2016scribblesup}, points \citep{bearman2016s}, and bounding boxes \citep{dai2015boxsup, khoreva2017simple, xu2015learning}. In case of image-level labels, class activation maps (CAMs) \citep{zhou2016learning} are used to localize the small discriminative regions which are not suitable particularly for the large-scale objects, but can be utilized as initial seeds (pseudo-masks) \citep{araslanov2020single, fan2020learning, sun2021ecs, kweon2021unlocking}.

\subsection{Zero-/Few-Shot Learning}
Motivated by humans' ability to recognize new concepts in a scene by using only a few visual samples, zero-shot and/or few-shot learning methods have been introduced. Few-shot semantic segmentation (FS3) methods \citep{wang2019panet, xie2021scale} has been proposed to recognize objects from unseen classes by utilizing few annotated examples; however, these methods are limited to handling a single unseen class only. Zero-shot semantic segmentation (ZS3) methods have been developed recently to generate visual features by exploiting word embedding vectors in the case of zero training samples \citep{bucher2019zero, xian2019semantic, pastore2021closer, lu2021feature}. However, the major drawback of ZS3 methods is their insufficient prediction ability to distinguish between the seen and the unseen classes even if both are included in a scene. This disadvantage is usually overcome by generalized ZS3 (GZS3), which recognizes both seen and unseen classes simultaneously. GZS3 studies mainly rely on generative-based methods. Feature extractor training is realized without considering semantic features in GZS3 adopted with generative approaches so that the bias is introduced towards the seen classes. Therefore, GZS3 methods result in performance reduction on unseen classes \citep{pastore2021closer}. Much of the recent work on ZS3 has involved such as; exploiting joint embedding space to alleviate the seen bias problem \citep{baek2021exploiting}, analyzing different domain performances \citep{ chan2021comprehensive}, and incorporating spatial information \citep{cheng2021sign}.

\subsection{Domain Adaptation}
Recent studies also rely on the use of synthetic large-scale image sets such as GTA5 \citep{richter2016playing} and SYNTHIA \citep{ros2016synthia} because of their capability to cope with laborious pixel-level annotations. Although these rich-labeled synthetic images have the advantage of reducing the labeling cost, they also bring about domain shift while training with unlabeled real images. Therefore, applying domain adaptation for aligning the synthetic and the real image sets is of much importance \citep{zhao2019multi, kang2020pixel, wu2021dannet, wang2021uncertainty, shin2021labor, fleuret2021uncertainty}. Unsupervised domain adaptation (UDA) methods are widely employed in semantic segmentation \citep{cheng2021dual, liu2021source, hong2018conditional, vu2019advent, pan2020unsupervised, wang2021domain, saporta2021multi, zheng2021rectifying}.

\subsection{Real-Time Processing}
Adopting compact and shallow model architectures \citep{zhao2018icnet, orsic2019defense, yu2018bisenet, li2019dfanet, fan2021rethinking} and restricting the input to be low-resolution \citep{marin2019efficient} are brand new innovations proposed very recently to overcome the computational burden of large-scale semantic segmentation. To choose a real-time semantic segmentation strategy, all aspects of an application should be considered, as all of these strategies somehow correlate with decreasing the model’s discriminative ability and losing information of object boundaries or small objects to some extent. Some other strategies have also been proposed for the retrieval of rich contextual information in real-time applications including attention mechanisms \citep{ding2021scarf, hu2020real}, depth-wise separable convolutions \citep{chollet2017xception, Howard2019}, pyramid fusion \citep{rosas2021fast, orvsic2021efficient}, grouped convolutions \citep{zhang2018shufflenet, huang2018condensenet} and neural architecture search \citep{zoph2018learning}, pipeline parallelism \citep{CHEW2022103995}.

\subsection{Contextual Information}
Contextual information aggregation with the purpose of augmenting pixel representations in semantic segmentation architectures is another promising research direction in recent years. In this aspect, mining contextual information \citep{jin2021mining}, exploring context information on spatial and channel dimensions \citep{li2021ctnet}, focusing on object based contextual representations \citep{yuan2020object} and capturing the global contextual information for fine-resolution remote sensing imagery \citep{li2021abcnet} are some of the recent studies. Alternative methods of reducing dense pixel-level annotations in semantic segmentation have been described which are based on using pixel-wise contrastive loss \citep{chaitanya2020contrastive, zhao2021contrastive, zhang2021looking}.

\section{Conclusions}
In this survey, we aimed at reviewing the current developments in the literature regarding deep learning-based 2D image semantic segmentation. We commenced with an analysis of the public image sets and leaderboards for 2D semantic segmentation and then continued by providing an overview of the techniques for performance evaluation. Following this introduction, our focus shifted to the 10-year evolution seen in this field under three chronological titles, namely the pre- and early- deep learning era, the fully convolutional era, and the post-FCN era. After a technical analysis on the approaches of each period, we presented a table of methods spanning all three eras, with a brief summary of each technique that explicates their contribution to the field.

In our review, we paid particular attention to the key technical challenges of the 2D semantic segmentation problem, the deep learning-based solutions that were proposed, and how these solutions evolved as they shaped the advancements in the field. To this end, we observed that the fine-grained localisation of pixel labels is clearly the definitive challenge to the overall problem. Although the title may imply a more ‘local’ interest, the research published in this field evidently shows that it is the global context that determines the actual performance of a method. Thus, it is eminently conceivable why the literature is rich with approaches that attempt to bridge local information with a more global context, such as graphical models, context aggregating networks, recurrent approaches, and attention-based modules. It is also clear that efforts to fulfil this local-global semantics gap at the pixel level will continue for the foreseeable future.

Another important revelation from this review has been the profound effect seen from public challenges to the field. Academic and industrial groups alike are in a constant struggle to top these public leaderboards, which has an obvious effect of accelerating development in this field. Therefore, it would be prudent to promote or even contribute to creating similar public image sets and challenges affiliated to more specific subjects of the semantic segmentation problem, such as 2D medical images.

Considering the rapid and continuing development seen in this field, there is an irrefutable need for an update on the surveys regarding the semantic segmentation problem. However, we believe that the current survey may be considered as a milestone in measuring how much the field has progressed thus far, and where the future directions possibly lie.

\bibliographystyle{chicagoB}  
\bibliography{ref}  

\end{document}